\documentclass[10pt,twocolumn,letterpaper]{article}

\usepackage[pagenumbers]{cvpr} 

\usepackage{booktabs}     
\usepackage{multirow}     
\usepackage{rotating}     
\usepackage{caption}      
\usepackage{indentfirst}
\usepackage[dvipsnames, svgnames]{xcolor} 
\definecolor{cvprblue}{rgb}{0.21,0.49,0.74}
\usepackage[pagebackref,breaklinks,colorlinks,allcolors=cvprblue]{hyperref}

\title{DocPTBench: Benchmarking End-to-End Photographed Document Parsing and Translation}

\author{
  Yongkun Du\textsuperscript{\rm 1,}$^*$, Pinxuan Chen\textsuperscript{\rm 1,}$^*$, Xuye Ying\textsuperscript{\rm 1,}$^*$, Zhineng Chen\textsuperscript{\rm 2,}$^\dag$
  \\[0.5em] 
   \textsuperscript{\rm 1}College of Computer Science and Artificial Intelligence, Fudan University, China\\
  \textsuperscript{\rm 2}Institute of Trustworthy Embodied AI, Fudan University, China\\[0.5em] 
  \texttt{\small \{ykdu23, pxchen25, xyying23\}@m.fudan.edu.cn, zhinchen@fudan.edu.cn}
}

\begin{document}
\maketitle
\renewcommand{\thefootnote}{}
\footnotetext{$^*$ The authors contributed equally.}
\footnotetext{$^\dag$ Corresponding author.}
\renewcommand{\thefootnote}{\arabic{footnote}}
\begin{abstract}

The advent of Multimodal Large Language Models (MLLMs) has unlocked the potential for end-to-end document parsing and translation. However, prevailing benchmarks such as OmniDocBench and DITrans are dominated by pristine scanned or digital-born documents, and thus fail to adequately represent the intricate challenges of real-world capture conditions, such as geometric distortions and photometric variations. To fill this gap, we introduce DocPTBench, a comprehensive benchmark specifically designed for Photographed Document Parsing and Translation. DocPTBench comprises over 1,300 high-resolution photographed documents from multiple domains, includes eight translation scenarios, and provides meticulously human-verified annotations for both parsing and translation. Our experiments demonstrate that transitioning from digital-born to photographed documents results in a substantial performance decline: popular MLLMs exhibit an average accuracy drop of 18\% in end-to-end parsing and 12\% in translation, while specialized document parsing models show significant average decrease of 25\%. This substantial performance gap underscores the unique challenges posed by documents captured in real-world conditions and reveals the limited robustness of existing models. Dataset and code are available at \url{https://github.com/Topdu/DocPTBench}.

\end{abstract}    
\section{Introduction}
\label{sec:intro}

Accurately parsing and translating documents~\cite{wang2024mineru,MinerU2,wei2024general,wei2024vary,feng2025dolphin,blecher2023nougat,cui2025paddleocr,liang2024document,zhang2023layoutdit,liang2025single,liang2025improving} from photographed images is fundamental to modern applications such as mobile scanning, multilingual invoice processing, and cross-lingual information retrieval. In these real-world scenarios, documents are often captured under uncontrolled conditions, resulting in images degraded by artifacts like motion blur, perspective distortion, and uneven lighting. Recent progress in Multimodal Large Language Models (MLLMs)~\cite{liu2023visual,achiam2023gpt,qwen2,InternVL,comanici2025gemini25pushingfrontier,yao2024minicpm,team2025kwai} offers a promising paradigm that unifies document parsing and translation into a single, end-to-end framework. By demonstrating strong performance on digital-born documents, these models have established the feasibility of this unified approach, challenging the traditional reliance on complex, multi-stage pipelines. Traditionally, this task relies on a sequence of discrete steps: document unwarping~\cite{ma2018docunet,verhoeven2023uvdoc,das2019dewarpnet,das2021end,ma2022learning} to rectify geometric distortions, followed by document-specialized models~\cite{wei2024general,cui2025paddleocr,wei2025deepseek,li2025monkeyocr,OCRFlux2025,mistral,Nanonets-OCR-S,dotsocr,chen2025ocean,liu2025points,chen2025dianjin,feng2025dolphin,vik2024marker,nassar2025smoldocling,du2024svtrv2,du2024igtr,su2025lranetpp}  for content extraction, and finally, translation of the resulting plain text. While MLLMs offer a compelling alternative to this cumbersome pipeline, their robustness on real-world photographed documents remains largely unverified due to a lack of appropriate benchmarks.

\begin{figure*}
  \centering
\includegraphics[width=0.98\textwidth]{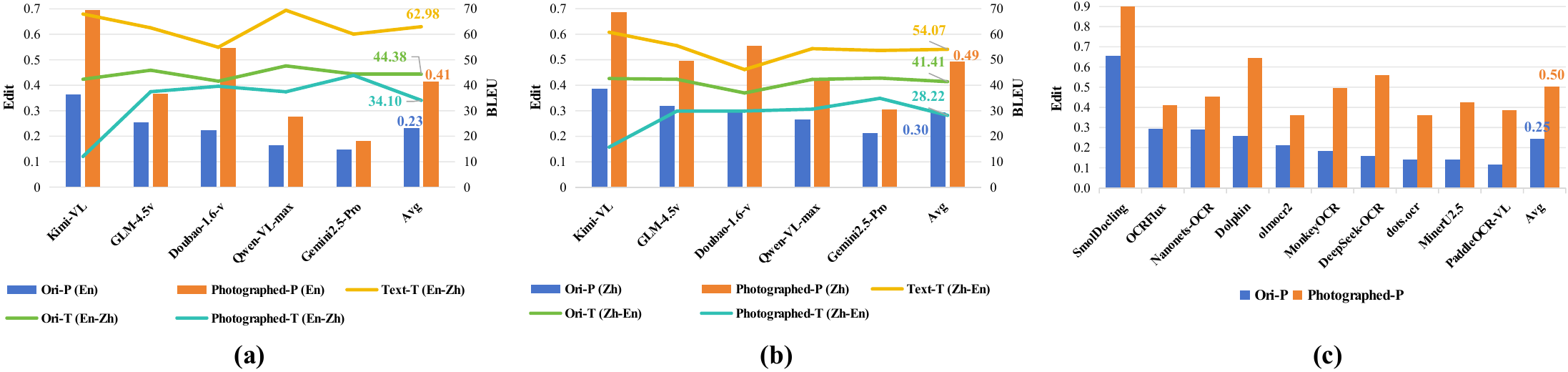} 

    \caption{\textbf{(a)}: the results of MLLMs on English (En)-started parsing (\textbf{P}) and translation (\textbf{T}) tasks; \textbf{(b)}: the counterpart on Chinese (Zh)-started tasks; \textbf{(c)}: the results from document parsing expert models. \textbf{Ori-} refers to the original digital-born document and \textbf{Photographed-} is its photographed version. \textbf{Text-} indicates that only the textual content of the document image is used as the source-language input. A lower \textbf{Edit} distance indicates higher parsing quality, and a higher \textbf{BLEU} score reflects better translation fidelity.
}
  \label{fig:fig1}
\end{figure*}

Prevailing benchmarks for document parsing and translation primarily evaluate model performance on digital-born or high-quality scanned documents.  Prominent parsing benchmarks like FoxPage~\cite{liu2024focus}, OmniDocBench~\cite{ouyang2025omnidocbench}, and olmOCR-Bench~\cite{poznanski2025olmocr} feature clean layouts and well-aligned text, failing to represent the complexities of photographed documents. Similarly, DoTA~\cite{liang2024document}, a pioneering benchmark for document image translation, is constructed from pristine LaTeX source files. Consequently, these benchmarks largely neglect the challenges inherent in photographed documents. While recent efforts like WildDoc~\cite{wang2025wilddoc} have begun to incorporate in-the-wild images, they focus on Visual Question Answering (VQA) and lack the parsing and translation annotations.

To bridge this gap, we introduce DocPTBench, a comprehensive benchmark for Photographed Document Parsing and Translation. DocPTBench comprises over 1,300 high-resolution document images spanning multiple domains, including invoices, forms, academic papers, and magazines, covering eight translation directions. Each image is paired with meticulously human-verified annotations for both parsing and translation. By integrating photographed images with a dual-task evaluation protocol, DocPTBench establishes a unified framework to comprehensively assess the parsing and translation capabilities of modern models. This design enables a direct and fair comparison between general-purpose MLLMs and specialized document parsing systems, a capability that was previously unattainable.

Our comprehensive evaluation on DocPTBench reveals a significant weakness in current models. Both state-of-the-art MLLMs and specialized document parsing models suffer substantial performance degradation when processing photographed documents. As illustrated in Fig.~\ref{fig:fig1}, this transition increases the average parsing edit distance by 18\% for MLLMs and 25\% for expert models. The results on translation is also severe, with BLEU scores dropping by 12\% when shifting to photographed documents. These findings expose key bottlenecks of modern models: a vulnerability to visual artifacts (e.g., noise, layout interference) and persistent challenges in spatial reasoning, low-light robustness, and multilingual typography handling. By providing a rigorous and realistic testbed, we aim to catalyze the development of more robust and generalizable document intelligence models, ultimately bridging the gap between laboratory performance and real-world applicability.

Our main contributions are summarized as follows:

\begin{itemize}
    \item We introduce DocPTBench, the first benchmark for real-world photographed document parsing and translation. It covers instances from multiple domains and languages and incorporates photo-taking conditions, aiming to reveal the practical challenges inherent to this task.
    \item We conduct extensive evaluations of both general-purpose MLLMs and document-parsing specialized models on DocPTBench. The results demonstrate a significant performance degradation when models are tasked with parsing or translating photographed documents, underscoring current limitations and identifying important directions for future research.
\end{itemize}

\section{Related Work}
\label{sec:relatedwork}

\noindent\textbf{Document Parsing Benchmarks.} The rapid evolution of vision-language models~\cite{bai2025qwen2,zhu2025internvl3,wang2025internvl3,comanici2025gemini25pushingfrontier,achiam2023gpt,team2025kimi,vteam2025glm45vglm41vthinkingversatilemultimodal,yang2025kwai,guo2025seed1} has made document parsing a cornerstone capability, spurring the creation of specialized benchmarks. Early efforts, such as FoxPage~\cite{liu2024focus}, concentrated on the structured domain of academic papers sourced from clean PDF documents. Subsequent work broadened this scope: OmniDocBench~\cite{ouyang2025omnidocbench} included multiple domains like digital-born PDFs and handwritten notes, while olmOCR-Bench~\cite{poznanski2025olmocr} was introduced to assess content recall. 

However, a common thread unites these benchmarks: a reliance on pristine, digital-born or high-quality scanned documents. Their content is typically well-structured and geometrically regular, failing to capture the complexities of real-world acquisition. The performance degradation on photographed documents has been noted in related tasks like MTVQA~\cite{tang2025mtvqa}, where MLLMs show a marked drop in accuracy. This observation prompted the creation of WildDoc~\cite{wang2025wilddoc}, a benchmark featuring in-the-wild document images. Despite its use of realistic images, WildDoc is designed as a document VQA~\cite{mathew2021docvqa,masry2022chartqa,kim2024tablevqa} dataset and thus lacks the detailed structural annotations required for assessing parsing systems. Consequently, a benchmark for parsing photographed documents with rich layout and content annotations remains a critical need, one that DocPTBench is designed to fill.

\noindent\textbf{Document Translation Benchmarks.} The evaluation of end-to-end document translation from images is an emerging research area. Foundational work includes DoTA~\cite{liang2024document}, which pioneered the task by translating document images (primarily from arXiv) into structured Markdown. Building on this, DITrans~\cite{zhang2023layoutdit} emphasized high-quality plain-text translation from documents with complex layouts. While its manually curated reading order annotations were a key contribution, the dataset consists mostly of scanned images and does not evaluate document parsing.

Other efforts have explored different facets of the problem. M3T~\cite{hsu2024m3t} was designed to assess whether VLMs could exploit visual context to improve translation (e.g., correcting errors using visual cues), rather than evaluating a full end-to-end parsing and translation pipeline. Similarly, DIT700K~\cite{zhang2025chaotic} is a large-scale dataset generated via an automated pipeline, making it valuable for model training but less suited for fine-grained benchmarking; moreover, it lacks photographed documents. While these prior efforts laid critical groundwork, they collectively reveal a persistent gap. A comprehensive benchmark for evaluating end-to-end structured document translation from real-world photographed images, across multiple domains and language pairs, has been conspicuously absent. DocPTBench addresses this void by unifying these challenges within a single, integrated evaluation framework.

\section{DocPTBench}

\begin{table*}[t]\footnotesize
\centering
\begin{tabular}{c|c|c|c|c}
\toprule
Dataset      & Task                & Type              & Source-Target                                                           & \#Image \\
\midrule
FoxPage~\cite{liu2024focus}      & Parsing             & Digital-Born        &   -                                                                      & 212     \\
OmniDocBench~\cite{ouyang2025omnidocbench} & Parsing             & Digital-Born        &   -                                                                      & 981     \\
olmOCR-Bench~\cite{poznanski2025olmocr} & Parsing             & Digital-Born        &   -                                                                      &  1,403       \\
DoTA~\cite{liang2024document}         & Translation         & Digital-Born        & En-Zh                                                                   & 1,003   \\
DIT700K~\cite{zhang2025chaotic}      & Translation         & Digital-Born        & En-Zh/De, Zh-En                                                         & 1,280   \\
DITrans~\cite{zhang2023layoutdit}      & Translation         & Digital-Born/Photographed & En-Zh                                                                   & 335/30     \\
\midrule
Ours         & Parsing+Translation & Digital-Born/Photographed/Unwarping & \begin{tabular}[c]{@{}c@{}}En-Zh/De/Fr/Ru\\ Zh-En/De/Fr/Ru\end{tabular} & 981/1381/1381   \\
\bottomrule
\end{tabular}
\caption{Comparisons between DocPTBench and existing benchmarks. Source-Target means the translation pairs.}
\label{tab:relatedwork}
\end{table*}

This section presents the DocPTBench evaluation framework, designed to rigorously evaluate MLLMs on real-world document parsing and translation tasks. We will elaborate on four key aspects: (1) the meticulous construction of the benchmark, including Original, Photographed, and Unwarping sets; (2) the suite of metrics for both parsing and translation; (3) the diverse range of models evaluated; and (4) the detailed experimental setup.

\subsection{Benchmark construction}

\begin{figure}[t]
  \centering

  \includegraphics[
      width=\columnwidth, 
      trim=0mm 13mm 0mm 13mm, 
      clip,                 
  ]{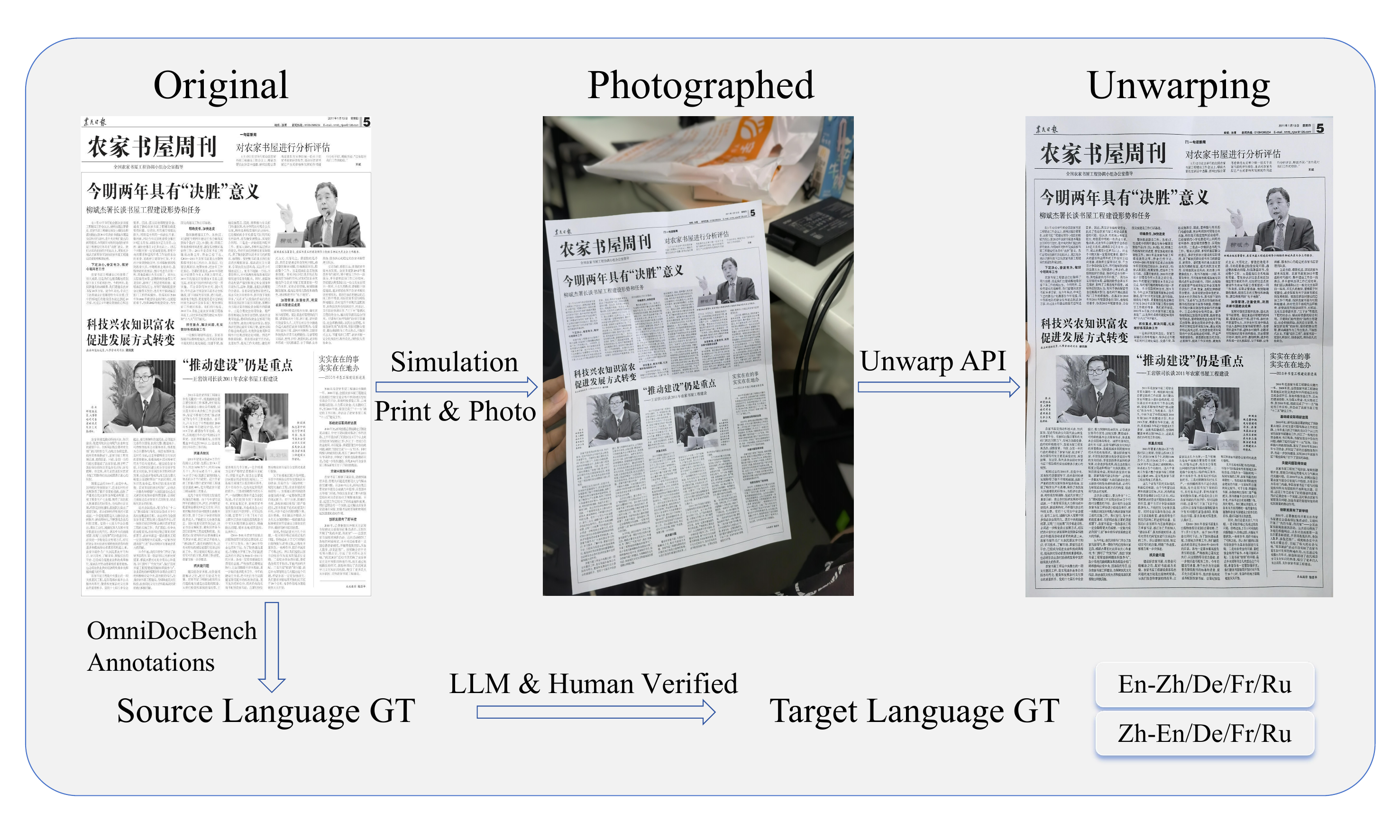} 
  \caption{
    Overview of the DocPTBench benchmark construction. 
  }
  \label{fig:data_pipeline}
\end{figure}

DocPTBench is meticulously designed to evaluate document parsing and translation models under realistic conditions. It is built upon the 981 high-quality documents from OmniDocBench~\cite{ouyang2025omnidocbench} and organized into a three-tiered structure for progressive, fine-grained analysis, as illustrated in Fig.~\ref{fig:data_pipeline}:

\textbf{Original}: This tier comprises the 981 original, born-digital documents. It serves as a  baseline to establish the upper-bound performance of a model's core parsing and translation capabilities, free from any visual degradation.

\textbf{Photographed}: To simulate real-world document acquisition, this tier contains 1,381 images. It includes: (1) 981 images generated by applying common photographic simulations to the original documents, and (2) 400 additional images created by physically photographing 100 of the original documents under four distinct and challenging conditions. This process introduces a diverse range of realistic degradations, including lighting variations (e.g., strong illumination, shadows), perspective distortions, physical wear (e.g., folds, wrinkles), and camera-induced artifacts like overexposure and motion blur.

\textbf{Unwarping}: This tier is created by processing the entire Photographed collection with a commercial document unwarping API\footnote{\url{https://www.textin.com/market/detail/crop\_enhance\_image}}. By rectifying geometric and curvature distortions, this set allows us to isolate their impact on model performance and to measure the effectiveness of such preprocessing pipelines.

To support the document translation task, we established ground truth for eight language pairs, including En-Zh/De/Fr/Ru and Zh-En/De/Fr/Ru. The source texts were taken from the human-verified annotations of OmniDocBench. For the target languages, we employed a semi-automated methodology: translations were first generated using the advanced Qwen-Max\footnote{\url{https://help.aliyun.com/zh/model-studio/models\#qwen-max-cn-bj}} and subsequently underwent rigorous manual verification and refinement. This approach ensures the creation of high-quality multilingual ground truth in a scalable and cost-effective manner.

As summarized in Tab.~\ref{tab:relatedwork}, DocPTBench distinguishes itself from existing benchmarks through three key features. It unifies the joint evaluation of parsing and translation, overcoming the limitation of prior work focusing on a single task. Furthermore, it incorporates 1,381 photographed images to assess model robustness under complex real-world visual conditions. Finally, its extensive multilingual support across eight language pairs significantly broadens the research scope. Together, these features establish DocPTBench as a critical resource for assessing the true end-to-end capabilities of document intelligence models in practical scenarios.

\subsection{Evaluation Metrics}
We evaluate model performance on two primary tasks: document parsing and document translation. Each task is assessed using a specialized suite of metrics for a comprehensive analysis. 

For document parsing, we adopt the evaluation suite from OmniDocBench~\cite{ouyang2025omnidocbench}, which assesses a model's ability to interpret various structural elements. These metrics include several based on Levenshtein edit distance~\cite{lcvenshtcin1966binary}, such as Text Edit, Formula Edit, Table Edit, and Read Order Edit, where lower scores signify better performance. Additionally, the Tree-Edit-Distance-based Similarity (TEDS) score~\cite{teds} is employed for table structure recognition, where a higher score indicates superior performance. 

For document translation, we employ a suite of standard machine translation metrics to assess translation quality. This includes BLEU~\cite{bleu_score} for n-gram precision, chrF~\cite{popovic2015chrf} for character-level n-grams, METEOR~\cite{banerjee2005meteor} for alignment-based unigram matching with stemming and synonymy, and STEDS~\cite{teds} for measuring semantic and structural similarity in the Markdown output.

\subsection{Models Evaluated}
To provide a rigorous and representative evaluation, our study assesses a diverse set of state-of-the-art models, which are specifically stratified into two distinct categories.

The first category, Expert Models, comprises systems specifically engineered and optimized for the document parsing task. This group includes leading purpose-built systems such as PaddleOCR-VL~\cite{cui2025paddleocr}, MinerU2.5~\cite{niu2025mineru2}, dots.ocr~\cite{dotsocr}, MonkeyOCR~\cite{li2025monkeyocr}, Deepseek-OCR~\cite{wei2025deepseek}, and olmOCR2~\cite{poznanski2025olmocr}. Given their focused design on document content extraction, these models are evaluated exclusively on document parsing tasks.

The second category, General MLLMs, consists of large-scale, general-purpose multimodal models endowed with broad visual and linguistic reasoning capabilities. We benchmark a selection of prominent models, including Gemini 2.5-Pro~\cite{comanici2025gemini25pushingfrontier}, Qwen-VL-Max~\cite{Qwen-VL}, GLM-4.5v~\cite{vteam2025glm45vglm41vthinkingversatilemultimodal}, Kimi-VL~\cite{team2025kimi}, and Doubao-Seed-1.6-vision~\cite{guo2025seed1}, on both parsing and translation to assess their versatility. Additionally, for the translation task, we include four leading open-source, parameter-efficient models: Qwen3-VL-4B~\cite{qwen3vl}, Qwen2.5-VL-3B~\cite{bai2025qwen2}, InternVL3-2B~\cite{zhu2025internvl3}, and InternVL3.5-2B~\cite{wang2025internvl3}. This inclusion allows us to establish a performance baseline for more accessible systems and to identify their current limitations on complex translation tasks.

\subsection{Experimental Setup}
To systematically evaluate the performance on the document parsing and translation tasks, we test models under various image conditions and prompting strategies, enabling a comprehensive analysis of model's strength and robustness against real-world visual distortions.

For the document parsing task, models are benchmarked across three distinct scenarios designed to progressively isolate specific challenges: (1) Original digital documents, which establish a performance baseline in ideal conditions; (2) Photographed images, which assess model robustness against complex conditions like perspective distortions and uneven lighting; and (3) Unwarping images, which serve to evaluate the specific impact of geometric rectification.

For the document translation task, our evaluation proceeds in several stages. First, we establish a text-only baseline to measure the upper bound of each model’s translation capability, separate from any visual parsing errors. Next, we assess end-to-end performance on document images using two prompting strategies: Simple prompt, which instructs the model to perform a direct, single-step translation from the image to target-language text; and Chain-of-Thought (CoT) prompt~\cite{wei2022chain}, which guides the model to first parse the document into source-language Markdown and then translate this intermediate representation. This two-step process helps decouple perception from translation, allowing for a more granular analysis of error sources. This end-to-end translation evaluation is conducted on both Original and Photographed documents, allowing us to quantify the performance degradation caused by real-world distortions. This layered methodology enables a comprehensive analysis of each model's strengths and weaknesses, from its core linguistic proficiency to its resilience against real-world visual degradation.

\newcommand{\loss}[1]{\textcolor{red}{\tiny$_{\downarrow#1}$}}
\newcommand{\gain}[1]{\textcolor{ForestGreen}{\tiny$_{\uparrow#1}$}}

\begin{table*}[!t]
\centering
\captionsetup{font=small} 

\scriptsize 
\setlength{\tabcolsep}{1pt} 
\renewcommand{\arraystretch}{1.05} 

\begin{tabular*}{\textwidth}{@{\extracolsep{\fill}} l l l ll ll ll ll ll ll @{}}
\toprule
\multirow{2}{*}{\textbf{Type}} & \multirow{2}{*}{\textbf{Model}} & \multirow{2}{*}{\textbf{Scene}} & \multicolumn{2}{c}{\textbf{Overall$^{\mathbf{Edit}\downarrow}$}} & \multicolumn{2}{c}{\textbf{Text$^{\mathbf{Edit}\downarrow}$}} & \multicolumn{2}{c}{\textbf{Formula$^{\mathbf{Edit}\downarrow}$}} & \multicolumn{2}{c}{\textbf{Table$^{\mathbf{TEDS}\uparrow}$}} & \multicolumn{2}{c}{\textbf{Table$^{\mathbf{Edit}\downarrow}$}} & \multicolumn{2}{c}{\textbf{Read Order$^{\mathbf{Edit}\downarrow}$}} \\
\cmidrule(lr){4-5} \cmidrule(lr){6-7} \cmidrule(lr){8-9} \cmidrule(lr){10-11} \cmidrule(lr){12-13} \cmidrule(lr){14-15} 
& & & \textbf{En} & \textbf{Zh} & \textbf{En} & \textbf{Zh} & \textbf{En} & \textbf{Zh} & \textbf{En} & \textbf{Zh} & \textbf{En} & \textbf{Zh} & \textbf{En} & \textbf{Zh} \\
\midrule
\multirow{18}{*}{\rotatebox{90}{\textbf{Expert Models}}} & \multirow{3}{*}{PaddleOCR-VL~\cite{cui2025paddleocr}} &Original& 10.5 & 12.6 & 4.1\phantom{0} & 6.2\phantom{0} & 24.1 & 31.6 & 88.0 & 92.1 & 9.3\phantom{0} & 6.2\phantom{0} & 4.5\phantom{0} & 6.3\phantom{0} \\
& & Photographed & 37.5\loss{27.0} & 39.6\loss{27.0} & 29.5\loss{25.4} & 38.0\loss{31.8} & 47.1\loss{23.0} & 50.3\loss{18.7} & 54.3\loss{33.7} & 64.8\loss{27.3} & 44.4\loss{35.1} & 31.9\loss{25.7} & 29.1\loss{24.6} & 38.2\loss{31.9} \\
& & Unwarping & 15.9\gain{21.6} & 21.1\gain{18.5} & 10.0\gain{19.5} & 17.4\gain{20.6} & 30.1\gain{17.0} & 38.2\gain{12.1} & 82.5\gain{28.2} & 83.1\gain{18.3} & 14.1\gain{30.3} & 13.4\gain{18.5} & 9.1\phantom{0}\gain{20.0} & 15.4\gain{22.8} \\
\cmidrule{2-15}
& \multirow{3}{*}{MinerU2.5~\cite{niu2025mineru2}} &Original& 11.1 & 17.4 & 5.0 & 7.4 & 25.8 & 47.3 & 88.3 & 89.2 & 8.9 & 8.3 & 4.5 & 6.8 \\
& & Photographed & 37.3\loss{26.2} & 47.4\loss{30.0} & 37.0\loss{32.0} & 53.6\loss{46.2} & 44.3\loss{18.5} & 62.0\loss{14.7} & 54.9\loss{33.4} & 59.8\loss{29.4} & 38.9\loss{30.0} & 33.5\loss{25.2} & 29.0\loss{24.5} & 40.3\loss{33.5} \\
& & Unwarping & 17.3\gain{20.0} & 25.2\gain{22.2} & 13.1\gain{23.9} & 19.1\gain{34.5} & 31.9\gain{12.4} & 52.2\gain{9.8} & 79.2\gain{24.3} & 81.1\gain{21.3} & 15.7\gain{23.2} & 14.6\gain{18.9} & 8.3\phantom{0}\gain{20.7} & 15.0\gain{25.3} \\
\cmidrule{2-15} 
& \multirow{3}{*}{dots.ocr~\cite{dotsocr}} &Original& 12.5 & 16.0 & 3.2 & 6.6 & 32.9 & 41.6 & 88.6 & 89.0 & 9.9 & 9.2 & 4.0 & 6.7 \\
& & Photographed & 33.7\loss{21.2} & 37.3\loss{21.3} & 29.8\loss{26.6} & 35.8\loss{29.2} & 39.2\loss{6.3} & 54.4\loss{12.8} & 63.7\loss{24.9} & 67.6\loss{21.4} & 33.0\loss{23.1} & 27.1\loss{17.9} & 32.8\loss{28.8} & 31.8\loss{25.1} \\
& & Unwarping & 16.3\gain{17.4} & 24.1\gain{13.2} & 8.3\phantom{0}\gain{21.5} & 20.9\gain{14.9} & 32.2\gain{7.0} & 42.0\gain{12.4} & 80.2\gain{16.5} & 82.3\gain{14.7} & 16.9\gain{16.1} & 14.6\gain{12.5} & 7.9\phantom{0}\gain{24.9} & 18.9\gain{12.9} \\
\cmidrule{2-15}
& \multirow{3}{*}{MonkeyOCR~\cite{li2025monkeyocr}} &Original& 14.6 & 22.1 & 6.8 & 11.8 & 27.2 & 45.2  & 81.3 & 85.5 & 14.9 & 13.4 & 9.3 & 17.9 \\
& & Photographed & 46.4\loss{31.8} & 52.8\loss{30.7} & 34.5\loss{27.7} & 43.9\loss{32.1} & 48.7\loss{21.5} & 61.6\loss{16.4} & 33.1\loss{48.2} & 37.4\loss{48.1} & 64.5\loss{49.6} & 61.5\loss{48.1} & 37.9\loss{28.6} & 44.1\loss{26.2} \\
& & Unwarping & 18.8\gain{27.6} & 31.9\gain{20.9} & 12.5\gain{22.0} & 23.6\gain{20.3} & 32.1\gain{16.6} & 55.8\gain{5.8} & 77.2\gain{44.1} & 77.1\gain{39.7} & 17.2\gain{47.3} & 19.5\gain{42.0} & 13.5\gain{24.4} & 28.7\gain{15.4} \\
\cmidrule{2-15}
& \multirow{3}{*}{DeepSeek-OCR~\cite{wei2025deepseek}} &Original& 13.4 & 18.1 & 4.6 & 9.7 & 28.5 & 43.3 & 82.6 & 89.0 & 13.8 & 8.8 & 6.7 & 10.5 \\
& & Photographed & 54.4\loss{41.0} & 57.8\loss{39.7} & 56.7\loss{52.1} & 57.6\loss{47.9} & 54.4\loss{25.9} & 74.1\loss{30.8} & 28.0\loss{54.6} & 35.4\loss{53.6} & 64.7\loss{50.9} & 59.2\loss{50.4} & 41.7\loss{35.0} & 40.4\loss{29.9} \\
& & Unwarping & 22.1\gain{32.3} & 33.5\gain{24.3} & 14.9\gain{41.8} & 29.4\gain{28.2} & 32.1\gain{22.3} & 58.8\gain{15.3} & 67.0\gain{39.0} & 75.8\gain{40.4} & 26.7\gain{38.0} & 20.9\gain{38.3} & 14.8\gain{26.9} & 24.9\gain{15.5} \\
\cmidrule{2-15}
& \multirow{3}{*}{olmOCR2~\cite{poznanski2025olmocr}} &Original& 16.1 & 26.7 & 4.8 & 18.5 & 39.2 & 54.3 & 83.7 & 78.5 & 12.3 & 16.5 & 8.1 & 17.4 \\
& & Photographed & 27.8\loss{11.7} & 44.6\loss{17.9} & 22.0\loss{17.2} & 39.9\loss{21.4} & 44.6\loss{5.4} & 74.1\loss{19.8} & 67.6\loss{16.1} & 65.4\loss{13.1} & 24.6\loss{12.3} & 28.5\loss{12.0} & 19.9\loss{11.8} & 36.0\loss{18.6} \\
& & Unwarping & 17.5\gain{10.3} & 37.2\gain{7.4} & 7.3\phantom{0}\gain{14.7} & 32.9\gain{7.0} & 37.5\gain{7.1} & 66.7\gain{7.4} & 81.9\gain{14.3} & 77.2\gain{11.8} & 14.3\gain{10.3} & 19.1\gain{9.4} & 11.0\gain{8.9} & 30.2\gain{5.8} \\
\midrule
\multirow{15}{*}{\rotatebox{90}{\textbf{General MLLMs}}} 
& \multirow{3}{*}{Gemini2.5-Pro \cite{comanici2025gemini25pushingfrontier}} &Original& 14.8 & 21.2 & 5.5 & 16.8 & 35.6 & 43.9  & 85.8 & 86.4 & 13.0 & 11.9 & 4.9 & 12.1 \\
& & Photographed & 18.2\loss{3.4} & 30.4\loss{9.2} & 9.8\phantom{0}\loss{4.3} & 27.7\loss{10.9} & 37.1\loss{1.5} & 56.8\loss{12.9} & 81.3\loss{4.5} & 82.9\loss{3.5} & 14.6\loss{1.6} & 13.7\loss{1.8} & 11.2\loss{6.3} & 23.6\loss{11.5} \\
& & Unwarping & 16.9\gain{1.3} & 27.3\gain{3.1} & 9.2\phantom{0}\gain{0.6} & 20.8\gain{6.9} & 35.3\gain{1.8} & 57.0\loss{0.2} & 83.4\gain{2.1} & 85.9\gain{3.0} & 13.1\gain{1.5} & 11.8\gain{1.9} & 10.0\gain{1.2} & 19.8\gain{3.8} \\
\cmidrule{2-15}
& \multirow{3}{*}{Doubao-1.6-v~\cite{guo2025seed1}} &Original& 22.5 & 29.3 & 16.2 & 27.6 & 31.2 & 47.2 & 66.6 & 76.3 & 31.9 & 24.5 & 10.8 & 17.9 \\
& & Photographed & 54.7\loss{32.2} & 55.4\loss{26.1} & 60.6\loss{44.4} & 58.2\loss{30.6} & 51.5\loss{20.3} & 61.1\loss{13.9} & 27.6\loss{39.0} & 37.9\loss{38.4} & 67.0\loss{35.1} & 61.9\loss{37.4} & 39.7\loss{28.9} & 40.2\loss{22.3} \\
& & Unwarping & 30.0\gain{24.7} & 42.5\gain{12.9} & 23.8\gain{36.8} & 41.8\gain{16.4} & 34.5\gain{17.0} & 56.4\gain{4.7} & 55.7\gain{28.1} & 60.8\gain{22.9} & 44.9\gain{22.1} & 42.4\gain{19.5} & 16.7\gain{23.0} & 29.5\gain{10.7} \\
\cmidrule{2-15}
& \multirow{3}{*}{Qwen-VL-Max~\cite{bai2025qwen2}} &Original& 16.6 & 26.5 & 5.2 & 20.5 & 32.9 & 44.0 & 84.2 & 86.7 & 22.0 & 23.7 & 6.5 & 17.7 \\
& & Photographed & 27.7\loss{11.1} & 42.7\loss{16.2} & 15.9\loss{10.7} & 41.5\loss{21.0} & 41.8\loss{8.9} & 57.2\loss{13.2} & 71.1\loss{13.1} & 71.6\loss{15.1} & 36.3\loss{14.3} & 38.0\loss{14.3} & 16.8\loss{10.3} & 34.4\loss{16.7} \\
& & Unwarping & 19.0\gain{8.7} & 32.6\gain{10.1} & 6.8\phantom{0}\gain{9.1} & 32.1\gain{9.4} & 33.8\gain{8.0} & 48.5\gain{8.7} & 81.3\gain{10.2} & 83.3\gain{11.7} & 26.5\gain{9.8} & 22.0\gain{16.0} & 9.0\phantom{0}\gain{7.8} & 27.8\gain{6.6} \\
\cmidrule{2-15}
& \multirow{3}{*}{GLM-4.5v~\cite{vteam2025glm45vglm41vthinkingversatilemultimodal}} &Original& 25.5 & 32.0 & 16.1 & 27.7 & 43.8 & 51.8 & 74.0 & 77.4 & 26.9 & 30.5 & 15.4 & 17.9 \\
& & Photographed & 36.7\loss{11.2} & 49.6\loss{17.6} & 26.2\loss{10.1} & 47.7\loss{20.0} & 49.9\loss{6.1} & 66.2\loss{14.4} & 58.9\loss{15.1} & 54.0\loss{23.4} & 43.5\loss{16.6} & 49.0\loss{18.5} & 27.3\loss{11.9} & 35.7\loss{17.8} \\
& & Unwarping & 23.9\gain{12.8} & 36.9\gain{12.7} & 13.1\gain{13.1} & 37.7\gain{10.0} & 39.0\gain{10.9} & 53.5\gain{12.7} & 73.8\gain{14.9} & 75.6\gain{21.6} & 26.9\gain{16.6} & 28.7\gain{20.3} & 16.5\gain{10.8} & 27.7\gain{8.0} \\
\cmidrule{2-15}
& \multirow{3}{*}{Kimi-VL~\cite{team2025kimi}} &Original& 36.5 & 38.7 & 17.2 & 22.0 & 48.6 & 52.2 & 57.1 & 67.8 & 65.9 & 62.5 & 14.3 & 18.1 \\
& & Photographed & 69.6\loss{33.1} & 68.7\loss{30.0} & 66.0\loss{48.8} & 63.5\loss{41.5} & 75.5\loss{26.9} & 82.6\loss{30.4} & 16.4\loss{40.7} & 22.9\loss{44.9} & 85.4\loss{19.5} & 82.2\loss{19.7} & 51.6\loss{37.3} & 46.7\loss{28.6} \\
& & Unwarping & 41.1\gain{28.5} & 50.7\gain{18.0} & 26.3\gain{39.7} & 38.5\gain{25.0} & 50.4\gain{25.1} & 68.8\gain{13.8} & 55.4\gain{39.0} & 62.3\gain{39.4} & 65.4\gain{20.0} & 65.0\gain{17.2} & 22.1\gain{29.5} & 30.7\gain{16.0} \\
\bottomrule
\end{tabular*}

\caption{Document parsing performance on DocPTBench across various image conditions. This table contrasts the accuracy of specialized Expert Models and General MLLMs on three document types: pristine Original documents, Photographed images with real-world distortions, and geometrically corrected Unwarping images. The results quantify the performance degradation caused by photographic artifacts and the partial recovery achieved through unwarping. Lower Edit scores and higher TEDS scores indicate better performance. }
\label{tab:parsing_results}

\end{table*}

\section{DocPTBench Results}
This section presents a diagnostic evaluation of model performance on DocPTBench, systematically analyzing the challenges inherent in real-world document parsing and translation. We first quantify the performance degradation caused by photographic distortions and the extent to which unwarping mitigates this loss. Then our analysis pivots to translation, dissecting the modality gap and the utility of Chain-of-Thought prompting. Finally, by correlating parsing accuracy with translation fidelity under various conditions, we pinpoint the primary bottlenecks for document translation, revealing a dual dependency on both visual perception robustness and intrinsic linguistic capability.

\subsection{Real-World Parsing Challenges}

The transition from digital-born documents to real-world photographed documents reveals a critical vulnerability across all evaluated models. As shown in Tab.~\ref{tab:parsing_results}, this shift introduces a stark and universal degradation in parsing accuracy. The introduction of the visual artifacts, ranging from geometric distortions like perspective distortions and curvature, to photometric variations such as uneven lighting and motion blur, severely compromises model performance.

This fragility is particularly pronounced in specialized expert models, which are often built on multi-stage pipelines. For instance, MonkeyOCR's Overall Edit Score (En) deteriorates dramatically from 14.6 on the Original set to 46.4 on the Photographed set. This significant decline is emblematic of a broader trend, with specialized expert models experiencing an average performance degradation of 25\%. This can be attributed to their architectural design, which often relies on layout analysis modules enforcing rectangular bounding box assumptions. Such constraints are fundamentally incompatible with the warped and distorted text regions common in photographed documents, leading to catastrophic failures in the parsing task.

In contrast, Gemini 2.5-Pro, one of the top-tier general-purpose MLLMs, demonstrates comparatively greater resilience, with the Overall Edit Score (En) moving from 14.8 to just 18.2. General-purpose MLLMs saw a more modest average decline of only 18\%, which suggests that their end-to-end architectures, likely trained on a more diverse and noisy dataset, are inherently better equipped to disentangle textual content from visual artifacts. This inherent robustness allows them to outperform many specialized expert models in these real-world conditions, even if their baseline performance on pristine documents is lower.

\subsection{Efficacy and Limitations of Unwarping}

To isolate the specific impact of geometric distortion, we conducted a further evaluation on the Unwarping dataset. The results, detailed in Tab.~\ref{tab:parsing_results}, confirm that this pre-processing step yields a substantial and system-wide performance recovery, establishing that restoring a document's geometric integrity is a cornerstone of accurate parsing. For instance, DeepSeek-OCR’s Overall Edit Score (En) improves dramatically from 54.4 on photographed images back to 22.1, recovering most of the performance lost from the pristine original. Similarly, the Table TEDS score for PaddleOCR-VL (En) rebounds from 54.3 to 82.5. This trend holds true across nearly all models and metrics, because unwarping rectifies perspective skew and page curvature, transforming distorted text regions back into the rectilinear layouts that most parsing models are fundamentally designed to process. This correction greatly reduces the burden on the model's layout analysis and OCR modules.

\begin{figure}[htbp]
  \centering
  \includegraphics[width=\columnwidth,
  trim={0cm 9mm 0cm 5mm}, 
  clip=true]
  {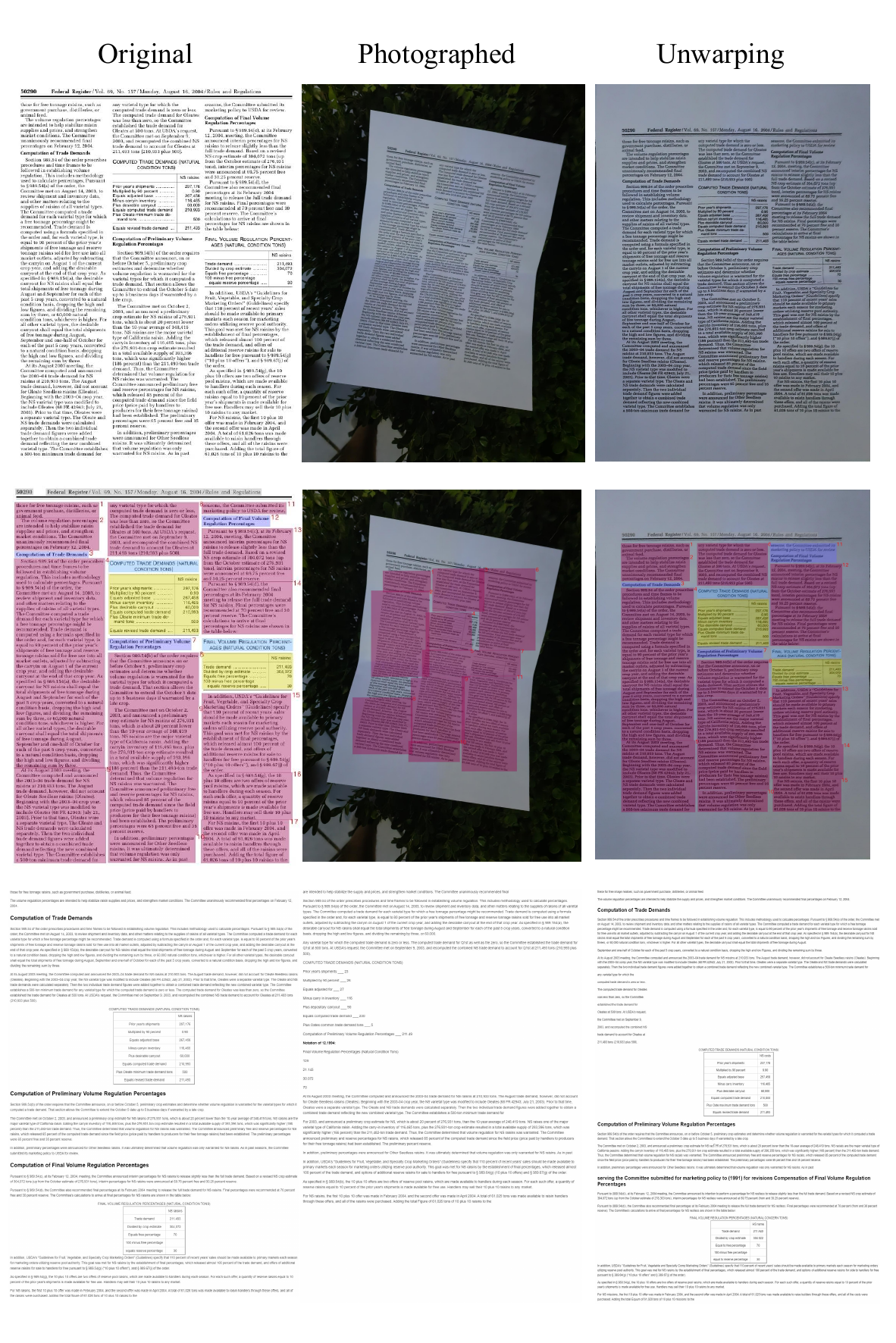} 
  \caption{
    Document parsing results of PaddleOCR-VL across different document conditions. The images visually corroborate our quantitative findings, demonstrating the sharp decline in parsing quality on photographed images and the significant improvement after unwarping. More cases are presented in the supplementary.
  }
  \label{fig:parsing_visualization}
\end{figure}

However, this recovery, while significant, is incomplete. A persistent performance gap remains between the results on the Unwarping set and the Original set. This lingering deficit empirically validates that geometric distortion is only part of the problem. Photometric degradations such as uneven lighting and motion blur, are not addressed by geometric correction and continue to impede model performance. These artifacts directly corrupt the low-level visual features processed by the vision encoder. As visually corroborated in Fig.~\ref{fig:parsing_visualization}, even after the document's macro-structure is perfectly flattened, the text itself may remain blurry compared to the digital-born one. While geometric correction is a crucial and effective pre-processing step, achieving true real-world robustness requires a two-pronged approach that combines better pre-processing with the development of model architectures that are intrinsically more resilient to a wide array of photometric noise.

\begin{table*}[!t]
\centering
\scriptsize 
\setlength{\tabcolsep}{4pt} 
\renewcommand{\arraystretch}{1.1} 
\begin{tabular}{c|c|c|llll|llll}
\toprule

\multirow{2}{*}{\textbf{Type}} & \multirow{2}{*}{\textbf{Model}} & \multirow{2}{*}{\textbf{Input}} & \multicolumn{4}{c}{\textbf{En-Zh}} & \multicolumn{4}{c}{\textbf{Zh-En}} \\
\cmidrule(lr){4-7}  \cmidrule(lr){8-11}
& & & \textbf{BLEU} & \textbf{chrF} & \textbf{METEOR} & \textbf{STEDS} & \textbf{BLEU} & \textbf{chrF} & \textbf{METEOR} & \textbf{STEDS} \\
\midrule
\multirow{12}{*}{\rotatebox{90}{\textbf{Open Source}}} & \multirow{3}{*}{Qwen3-VL-4B~\cite{qwen3vl}} & Text & 49.61 & 56.87 & 66.74 & 94.35 & 50.20 & 72.82 & 64.91 & 94.24 \\
& & Original-Simple & 32.11\loss{17.50} & 40.22\loss{16.65} & 47.49\loss{19.25} & 64.55\loss{29.80} & 28.31\loss{21.89} & 48.72\loss{24.10} & 40.44\loss{24.47} & 68.41\loss{25.83} \\
& & Original-CoT & 36.86\gain{4.75} & 45.17\gain{4.95} & 53.97\gain{6.48} & 68.83\gain{4.28} & 34.84\gain{6.53} & 57.29\gain{8.57} & 48.75\gain{8.31} & 66.14\loss{2.27} \\
\cmidrule{2-11}
& \multirow{3}{*}{Qwen2.5-VL-3B~\cite{bai2025qwen2}} & Text & 48.60 & 55.39 & 63.91 & 81.59 & 45.29 & 66.13 & 57.55 & 87.35 \\
& & Original-Simple & 18.18\loss{30.42} & 25.65\loss{29.74} & 27.42\loss{36.49} & 59.02\loss{22.57} & 15.20\loss{30.09} & 23.73\loss{42.40} & 20.78\loss{36.77} & 60.87\loss{26.48} \\
& & Original-CoT & 19.37\gain{1.19} & 28.85\gain{3.20} & 32.09\gain{4.67} & 49.57\loss{9.45} & 18.50\gain{3.30} & 35.56\gain{11.83} & 28.98\gain{8.20} & 48.24\loss{12.63} \\
\cmidrule{2-11}
& \multirow{3}{*}{InternVL3-2B~\cite{zhu2025internvl3}} & Text & 48.25 & 54.29 & 62.48 & 89.42 & 33.54 & 50.01 & 43.78 & 84.94 \\
& & Original-Simple & 10.87\loss{37.38} & 17.33\loss{36.96} & 18.91\loss{43.57} & 55.90\loss{33.52} & 7.27\phantom{0}\loss{26.27} & 11.63\loss{38.38} & 10.38\loss{33.40} & 57.83\loss{27.11} \\
& & Original-CoT & 19.21\gain{8.34} & 28.07\gain{10.74} & 32.91\gain{14.00} & 55.16\loss{0.74} & 22.07\gain{14.80} & 46.01\gain{34.38} & 36.06\gain{25.68} & 59.16\gain{1.33} \\
\cmidrule{2-11}
& \multirow{3}{*}{InternVL3.5-2B~\cite{wang2025internvl3}} & Text & 57.49 & 63.14 & 72.23 & 94.29 & 48.46 & 69.48 & 61.02 & 92.18 \\
& & Original-Simple & 25.43\loss{32.06} & 34.62\loss{28.52} & 40.15\loss{32.08} & 64.44\loss{29.85} & 8.42\phantom{0}\loss{40.04} & 11.04\loss{58.44} & 10.52\loss{50.50} & 65.03\loss{27.15} \\
& & Original-CoT & 31.42\gain{5.99} & 41.25\gain{6.63} & 48.69\gain{8.54} & 65.14\gain{0.70} & 28.28\gain{19.86} & 50.16\gain{39.12} & 41.75\gain{31.23} & 61.86\loss{3.17} \\
\midrule
\multirow{15}{*}{\rotatebox{90}{\textbf{Closed Source}}} & \multirow{3}{*}{Gemini2.5-Pro~\cite{comanici2025gemini25pushingfrontier}} & Text & 60.07 & 66.54 & 76.39 & 92.90 & 53.62 & 76.01 & 70.06 & 91.23 \\
& & Original-Simple & 44.34\loss{15.73} & 53.83\loss{12.71} & 64.97\loss{11.42} & 71.77\loss{21.13} & 37.96\loss{15.66} & 67.45\loss{8.56} & 58.04\loss{12.02} & 65.75\loss{25.48} \\
& & Original-CoT & 44.41\gain{0.07} & 53.94\gain{0.11} & 65.68\gain{0.71} & 75.05\gain{3.28} & 42.81\gain{4.85} & 69.62\gain{2.17} & 61.67\gain{3.63} & 75.37\gain{9.62} \\
\cmidrule{2-11}
& \multirow{3}{*}{Qwen-VL-Max~\cite{Qwen-VL}} & Text & 69.41 & 74.05 & 82.81 & 96.91 & 54.33 & 75.19 & 67.35 & 92.19 \\
& & Original-Simple & 41.04\loss{28.37} & 50.81\loss{23.24} & 59.77\loss{23.04} & 72.76\loss{24.15} & 36.29\loss{18.04} & 61.03\loss{14.16} & 50.40\loss{16.95} & 71.68\loss{20.51} \\
& & Original-CoT & 47.60\gain{6.56} & 55.70\gain{4.89} & 64.10\gain{4.33} & 72.67\loss{0.09} & 42.28\gain{5.99} & 66.05\gain{5.02} & 56.44\gain{6.04} & 69.68\loss{2.00} \\
\cmidrule{2-11}
& \multirow{3}{*}{GLM-4.5v~\cite{vteam2025glm45vglm41vthinkingversatilemultimodal}} & Text & 62.53 & 68.38 & 77.84 & 95.57 & 55.51 & 75.62 & 68.56 & 92.84 \\
& & Original-Simple & 42.14\loss{20.39} & 51.20\loss{17.18} & 60.82\loss{17.02} & 73.72\loss{21.85} & 39.02\loss{16.49} & 62.67\loss{12.95} & 53.10\loss{15.46} & 74.34\loss{18.50} \\
& & Original-CoT & 45.90\gain{3.76} & 55.09\gain{3.89} & 64.91\gain{4.09} & 73.14\loss{0.58} & 42.34\gain{3.32} & 66.92\gain{4.25} & 57.48\gain{4.38} & 72.43\loss{1.91} \\
\cmidrule{2-11}
& \multirow{3}{*}{Kimi-VL~\cite{team2025kimi}} & Text & 67.95 & 72.45 & 81.78 & 97.34 & 60.76 & 78.64 & 73.47 & 95.61 \\
& & Original-Simple & 38.20\loss{29.75} & 47.17\loss{25.28} & 55.14\loss{26.64} & 70.38\loss{26.96} & 32.07\loss{28.69} & 54.72\loss{23.92} & 44.93\loss{28.54} & 69.85\loss{25.76} \\
& & Original-CoT & 42.36\gain{4.16} & 50.94\gain{3.77} & 58.68\gain{3.54} & 68.66\loss{1.72} & 42.63\gain{10.56} & 64.24\gain{9.52} & 55.75\gain{10.82} & 69.03\loss{0.82} \\
\cmidrule{2-11}
& \multirow{3}{*}{Doubao-1.6-v~\cite{guo2025seed1}} & Text & 54.92 & 62.59 & 72.26 & 87.26 & 46.15 & 71.22 & 62.51 & 83.70 \\
& & Original-Simple & 39.29\loss{15.63} & 49.73\loss{12.86} & 59.29\loss{12.97} & 69.80\loss{17.46} & 34.31\loss{11.84} & 61.94\loss{9.28} & 51.50\loss{11.01} & 70.99\loss{12.71} \\
& & Original-CoT & 41.61\gain{2.32} & 51.09\gain{1.36} & 61.32\gain{2.03} & 71.52\gain{1.72} & 36.98\gain{2.67} & 64.47\gain{2.53} & 54.26\gain{2.76} & 71.98\gain{0.99} \\
\bottomrule
\end{tabular}

\caption{End-to-End translation performance on digital-born (Original) documents. We quantify the performance gap between a model's pure text translation capability (Text-only baseline) and its end-to-end performance on document images. It further compares a Simple prompting strategy against a CoT approach to evaluate how decoupling perception and translation affects accuracy. Results for the other six language pairs are detailed in the supplementary material.}
\label{tab:translation_result}

\end{table*}

\subsection{Modality Gap in Document Translation}
To evaluate end-to-end document translation, we begin by quantifying the performance gap between translating text-only inputs and translating from document images. While our benchmark covers eight language pairs, for the sake of brevity, our main analysis focuses on the English-Chinese (En-Zh) and Chinese-English (Zh-En) translation directions. Comprehensive results for all language pairs are provided in the supplementary material. The results, detailed in Tab.~\ref{tab:translation_result}, reveal a substantial and widespread performance degradation across all models, highlighting the inherent challenge of transitioning from a unimodal, linguistic task to a multimodal one. For instance, Qwen-VL-Max’s BLEU score on the En-Zh translation task plummets from 69.41 to 41.04. The performance collapse is even more pronounced for other models, such as InternVL3.5-2B on the Zh-En task, where the METEOR score falls from 61.02 to a mere 10.52, underscoring the severity of this modality gap.

This performance gap is attributed to the compounded challenges of the multimodal task, which introduces a cascade of potential errors, including (1) Optical Character Recognition (OCR) inaccuracies that introduce corruption into the recognized source text; (2) misinterpretation of complex layouts that disrupts the logical flow of content; and (3) interference from non-textual visual elements that distracts the model from the relevant textual content. Critically, our analysis reveals that this complexity can lead to a more fundamental type of failure: some models, particularly smaller open-source variants, may fail to adhere to the translation instruction altogether and revert to performing a simple OCR of the document, as illustrated in Fig.~\ref{fig:translation_visualization}(b). This inability to execute the primary directive demonstrates that the seamless fusion of visual perception and linguistic translation remains a formidable challenge for current MLLMs.

\subsection{Decoupling Perception and Translation via CoT}
\begin{figure}[htbp]
  \centering
    \includegraphics[width=\columnwidth,
  trim={0cm 30mm 0cm 18mm}, 
  clip=true]
  {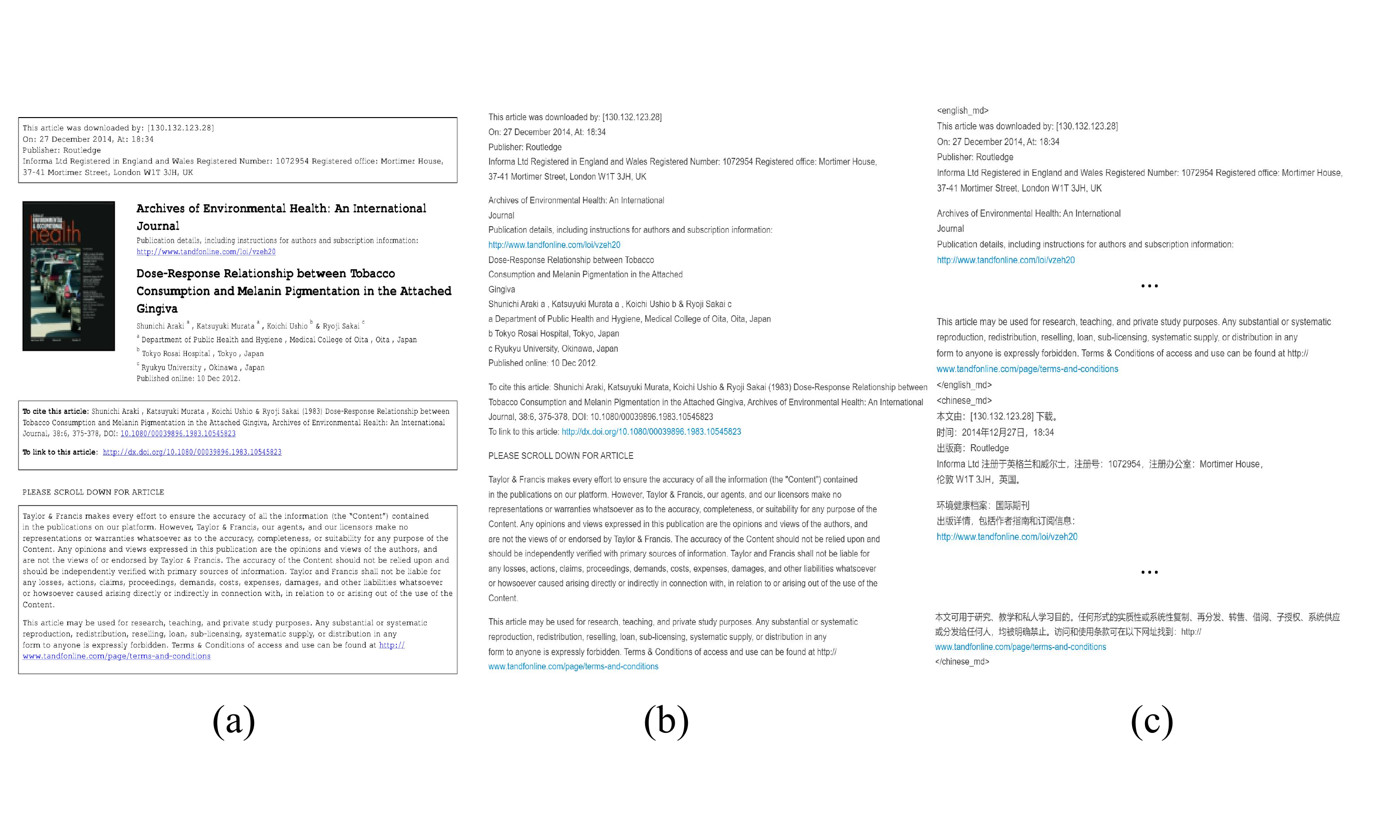}
  \caption{
    Visualization of Qwen3-VL-4B end-to-end translation results on DocPTBench.
    (a) shows the document image input. (b) illustrates a failure case where using a simple prompt causes the model to only perform OCR without translating. (c) demonstrates that the CoT prompt successfully rectifies this failure, guiding the model to produce the correct translation.
  }
  \label{fig:translation_visualization}
\end{figure}

To address the instruction-following failures observed with simple prompts, we implemented a Chain-of-Thought (CoT) prompting strategy. This approach improves instruction adherence by explicitly decoupling the complex task into two simpler, sequential steps (first extracting the text and then translating it) while preserving an end-to-end workflow. As illustrated in Fig.~\ref{fig:translation_visualization}(c), this method successfully corrects the aforementioned failure mode, guiding the model to produce the correct translation in cases where a simple prompt would fail. The data in Tab.~\ref{tab:translation_result} confirms the efficacy of this strategy, showing consistent performance recovery across all evaluated models and metrics. For instance, with the CoT strategy, GLM-4.5v’s En-Zh BLEU score improves from 42.14 to 45.90, and InternVL3-2B's Zh-En chrF score increases dramatically from 11.63 to 46.01. This significant improvement strongly suggests that many of the errors in the direct end-to-end approach stem from the model's difficulty in simultaneously performing recognition and translation.

However, the CoT strategy is not a panacea, as a considerable performance gap persists when compared to the text-only baseline, indicating a ceiling on its effectiveness. Beyond this performance ceiling, the CoT strategy also introduces practical limitations. First, its two-step instruction is less intuitive and does not align with how a typical user might naturally formulate a request. Second, this approach incurs significant token overhead, as the model generates the entire document's content in the source language as an intermediate step before providing the final translation.

\begin{table*}[!t]
\centering
\scriptsize 
\setlength{\tabcolsep}{4pt}
\renewcommand{\arraystretch}{1.1}

\begin{tabular}{c|c|llll|llll}
\toprule
\multirow{2}{*}{\textbf{Model}}                 & \multirow{2}{*}{\textbf{Input}} & \multicolumn{4}{c}{\textbf{En-Zh}}      & \multicolumn{4}{c}{\textbf{Zh-En}}      \\
\cmidrule(lr){3-6}  \cmidrule(lr){7-10}
                                        &                        & \textbf{BLEU}  & \textbf{chrF}  & \textbf{METEOR} & \textbf{STEDS} & \textbf{BLEU}  & \textbf{chrF}  & \textbf{METEOR} & \textbf{STEDS} \\
\midrule
\multirow{2}{*}{Gemini2.5-Pro~\cite{comanici2025gemini25pushingfrontier}} 
& Photographed-Simple & 43.72\loss{0.62} & 53.77\loss{0.06} & 63.68\loss{1.29} & 71.82\gain{0.05} & 32.88\loss{5.08} & 62.95\loss{4.50} & 52.24\loss{5.80} & 63.42\loss{2.33} \\
& Photographed-CoT    & 43.88\loss{0.53} & 53.88\loss{0.06} & 64.06\loss{1.62} & 75.18\gain{0.13} & 34.89\loss{7.92} & 61.59\loss{8.03} & 51.88\loss{9.79} & 70.26\loss{5.11} \\
\midrule
\multirow{2}{*}{Qwen-VL-Max~\cite{Qwen-VL}} 
& Photographed-Simple & 27.53\loss{13.51} & 37.25\loss{13.56} & 43.81\loss{15.96} & 69.02\loss{3.74} & 21.81\loss{14.48} & 45.93\loss{15.10} & 34.44\loss{15.96} & 64.96\loss{6.72} \\
& Photographed-CoT    & 37.44\loss{10.16} & 46.76\loss{8.94} & 54.99\loss{9.11} & 68.24\loss{4.43} & 30.64\loss{11.64} & 54.88\loss{11.17} & 44.43\loss{12.01} & 64.16\loss{5.52} \\
\midrule
\multirow{2}{*}{GLM-4.5v~\cite{vteam2025glm45vglm41vthinkingversatilemultimodal}} 
& Photographed-Simple & 31.03\loss{11.11} & 41.02\loss{10.18} & 47.41\loss{13.41} & 71.21\loss{2.51} & 24.82\loss{14.20} & 46.42\loss{16.25} & 37.45\loss{15.65} & 60.44\loss{13.90} \\
& Photographed-CoT    & 37.48\loss{8.42} & 46.72\loss{8.37} & 54.39\loss{10.52} & 70.94\loss{2.20} & 29.88\loss{12.46} & 53.71\loss{13.21} & 44.15\loss{13.33} & 62.60\loss{9.83} \\
\midrule
\multirow{2}{*}{Kimi-VL~\cite{team2025kimi}} 
& Photographed-Simple & 9.16\phantom{0}\loss{29.04} & 15.97\loss{31.20} & 20.51\loss{34.63} & 49.05\loss{21.33} & 9.15\phantom{0}\loss{22.92} & 27.77\loss{26.95} & 18.52\loss{26.41} & 50.99\loss{18.86} \\
& Photographed-CoT    & 12.07\loss{30.29} & 19.17\loss{31.77} & 23.46\loss{35.22} & 52.42\loss{16.24} & 15.78\loss{26.85} & 34.88\loss{29.36} & 26.49\loss{29.26} & 49.07\loss{19.96} \\
\midrule
\multirow{2}{*}{Doubao-1.6-v~\cite{guo2025seed1}} 
& Photographed-Simple & 35.36\loss{3.93} & 46.47\loss{3.26} & 53.60\loss{5.69} & 66.46\loss{3.34} & 26.88\loss{7.43} & 53.62\loss{8.32} & 42.58\loss{8.92} & 63.27\loss{7.72} \\
& Photographed-CoT    & 39.61\loss{2.00} & 49.61\loss{1.48} & 57.88\loss{3.44} & 66.70\loss{4.82} & 29.91\loss{7.07} & 56.52\loss{7.95} & 45.97\loss{8.29} & 63.53\loss{8.45} \\
\bottomrule
\end{tabular}
\caption{Impact of photographic distortions on end-to-end translation. We quantify the performance degradation caused by real-world visual distortions by directly comparing model performance on Photographed captures against the Original digital baseline. The subscript values indicate the performance change relative to the Original baseline, which is detailed in Tab.~\ref{tab:translation_result}.}
\label{tab:translation_photodrop}
\end{table*}

\subsection{Translation Collapse on Photographs}
The challenges of end-to-end translation are severely exacerbated when models process real-world photographed documents. As detailed in Tab.~\ref{tab:translation_photodrop}, the presence of authentic visual distortions, such as geometric warping and uneven lighting, causes an average decline of 12\% on BLEU score. This performance collapse is particularly significant because it occurs with both simple and CoT prompting methods, indicating that poor image quality, rather than the prompting strategy, is the primary bottleneck.

This degradation is pervasive across all evaluated models. For instance, under a CoT prompt, Qwen-VL-Max’s En-Zh BLEU score drops from 47.60 to 37.44, while Kimi-VL’s score plummets from 42.63 to a near-unusable 15.78 (Tab.~\ref{tab:translation_photodrop}). This failure mode is consistent with our earlier parsing analysis: visual artifacts severely degrade parsing quality, creating a corrupted and unreliable textual foundation for the subsequent translation task. This finding confirms that poor image quality is the fundamental bottleneck, compromising the robustness of these end-to-end systems and casting significant doubt on their viability for uncontrolled, real-world applications.

\subsection{Dual Bottlenecks for Document Translation}

Our final analysis reveals a strong positive correlation between an MLLM’s document parsing capabilities and its translation performance, as models with superior parsing abilities consistently deliver higher-quality translations. This dependency stems from the cascading nature of errors: inaccuracies in the parsing stage propagate and degrade the final translation. The challenge posed by photographed documents starkly exemplifies this phenomenon, as the visual artifacts in these images create a perceptual bottleneck, impeding the model's ability to accurately parse the document, which in turn degrades the quality of the translation.

Nevertheless, this relationship is also shaped by a model's intrinsic linguistic capabilities. The Doubao-Seed-1.6-vision model provides a compelling case study. Despite achieving strong document parsing scores, this model delivers only modest end-to-end translation quality. The reason for this discrepancy lies in its core translation engine; its performance on text-only translation tasks is significantly weaker than that of other leading models. This example demonstrates that end-to-end translation quality is governed by a dual bottleneck: the parsing accuracy of the vision component and the translation proficiency of the language component. Consequently, while substandard parsing inevitably compromises translation output, even flawless parsing cannot compensate for an inferior underlying translation engine.

\section{Conclusion}

This paper introduces DocPTBench, the first benchmark for end-to-end document parsing and translation on photographed images. It contains over 1,300 meticulously labeled images and aims to address a critical gap in evaluating document AI models under real-world conditions. DocPTBench enables the systematic and fair assessment of leading models against realistic visual distortions, providing crucial insights into their performance limitations beyond pristine, digital-born documents. By open-sourcing DocPTBench as a crucial resource for the community, we hope it will catalyze research into more resilient and accurate models for real-world scenarios, ultimately bridging the gap between laboratory performance and practical applicability.

{
    \small
    \bibliographystyle{ieeenat_fullname}
    \bibliography{main}
}

\clearpage
\setcounter{page}{1}
\maketitlesupplementary

\section{Overview}
In this supplementary material, we provide a comprehensive evaluation of the DocPTBench benchmark. Sec.~\ref{sec:supp_parsing} details the document parsing task, offering an in-depth quantitative analysis of the full model suite. Sec.~\ref{sec:supp_translation} extends the document translation analysis to six additional language pairs (En-De/Fr/Ru and Zh-De/Fr/Ru), with a specific focus on the modality gap and the impact of prompting strategies across all languages. We also offer qualitative visualization cases to facilitate a better understanding of our work.

\section{Document Parsing}
\label{sec:supp_parsing}
Tab.~\ref{tab:parsing_results_full} presents the complete benchmarking results for all 18 evaluated systems, including additional Expert Models (Dolphin~\cite{feng2025dolphin}, olmOCR~\cite{poznanski2025olmocr}, OCRFlux~\cite{OCRFlux2025}, SmolDocling~\cite{nassar2025smoldocling}, Nanonets~\cite{Nanonets-OCR-S}) and the large-scale General MLLM (Qwen2.5-VL-72B~\cite{bai2025qwen2}).

\subsection{Vulnerability to Photographic Distortion}
The expanded results reveal a severe lack of robustness across both specialized document parsing models and general MLLMs when facing real-world geometric distortions.  For instance, SmolDocling's Overall Edit Score (En) deteriorates catastrophically from 49.3 to 90.1, and Dolphin drops from 20.5 to 57.5. General MLLMs are also not immune to this trend. Despite their massive pre-training, models like Qwen2.5-VL-72B~\cite{bai2025qwen2} and Kimi-VL~\cite{team2025kimi} experience significant degradations, with their scores worsening by 20.1 and 33.1 points, respectively. While Gemini2.5-Pro~\cite{comanici2025gemini25pushingfrontier} shows relative resilience (dropping only 3.4 points), the widespread decline confirms that most current architectures, whether specialized for OCR or general-purpose, remain fundamentally brittle when processing curved or perspective-distorted text.

\subsection{Critical Role of Unwarping}
The results on the \textit{Unwarping} set provide strong empirical evidence for the necessity of geometric rectification. Across almost all expert models, unwarping significantly restores performance. Dolphin~\cite{feng2025dolphin} recovers its Overall Edit Score (En) from 57.5 to 27.3, and OCRFlux~\cite{OCRFlux2025} sees its Table TEDS score (En) rebound from 49.5 to 68.1. However, the recovery is rarely complete, and a performance gap remains compared to the \textit{Original} set (e.g., DeepSeek-OCR~\cite{wei2025deepseek} is still 8.7 points worse on Unwarping than Original). This residual gap indicates that while unwarping fixes layout issues, it cannot fully compensate for photometric degradations like blur and uneven lighting.

\subsection{Qualitative Visualizations}
Fig.~\ref{fig:parsing_visualization_1_paddle} to Fig.~\ref{fig:parsing_visualization_5_mineru} visualize the parsing outputs of different models under varying evaluation conditions. These examples corroborate our quantitative findings that expert models frequently miss entire text blocks or scramble the reading order due to page curvature.
Furthermore, flattening the image rectifies most layout errors, though artifacts like motion blur continue to cause character-level recognition errors.

\section{Document Translation}
\label{sec:supp_translation}
\begin{figure*}[!t]
  \centering
  \includegraphics[width=0.88\textwidth, clip, trim={0 5mm 0 0mm}]{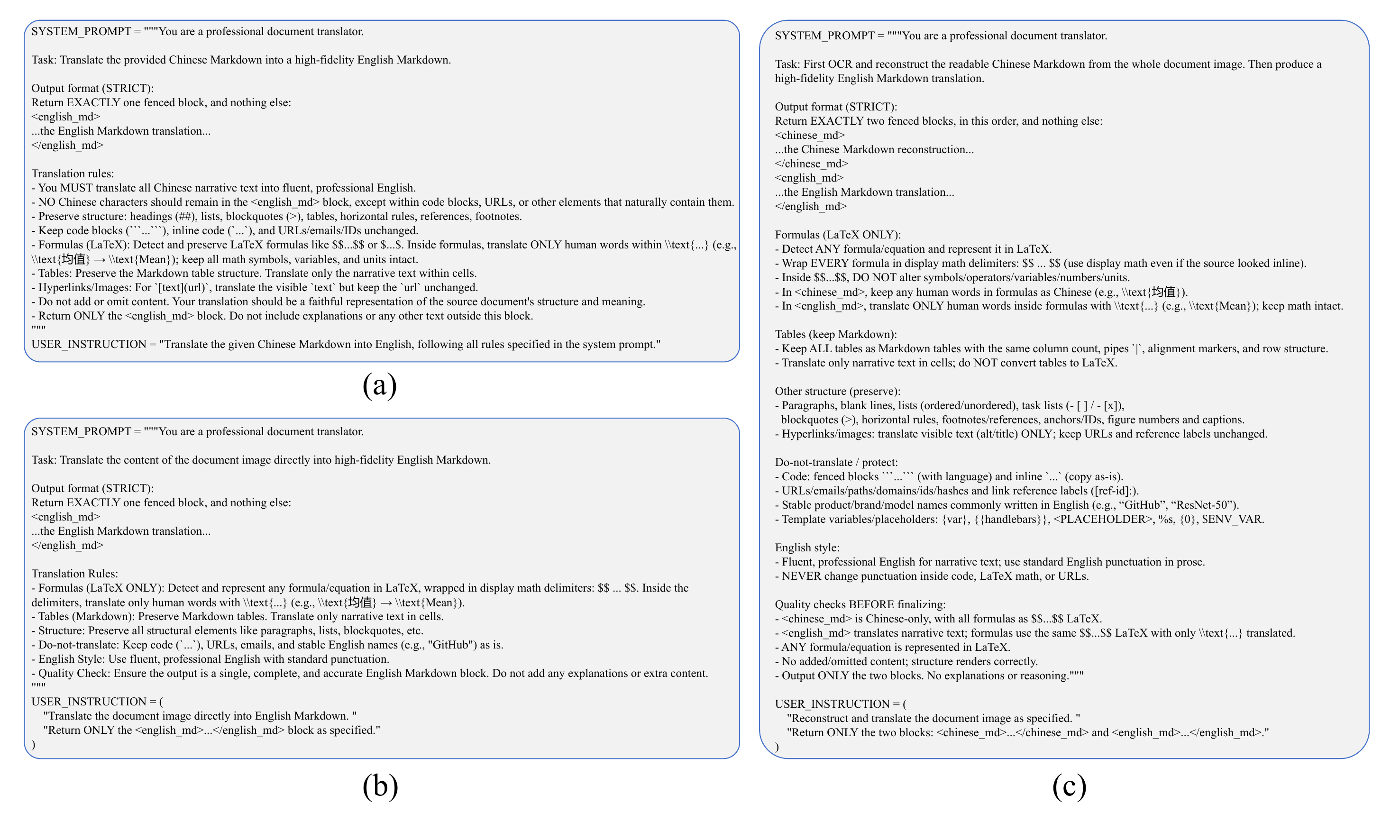}
  
  \caption{
    Illustration of the prompting strategies employed in our document translation experiments. (a) Text Prompt: Used for the text-only machine translation baseline. (b) Simple Prompt: Direct instruction for end-to-end visual document translation. (c) CoT Prompt: A Chain-of-Thought approach that explicitly instructs the model to perform OCR recognition prior to translation to mitigate modality gaps.
  }
  \label{fig:prompt}
\end{figure*}

\begin{figure*}[!t]
  \centering
  \includegraphics[width=0.88\textwidth, clip, trim={0 3mm 0 6mm}]{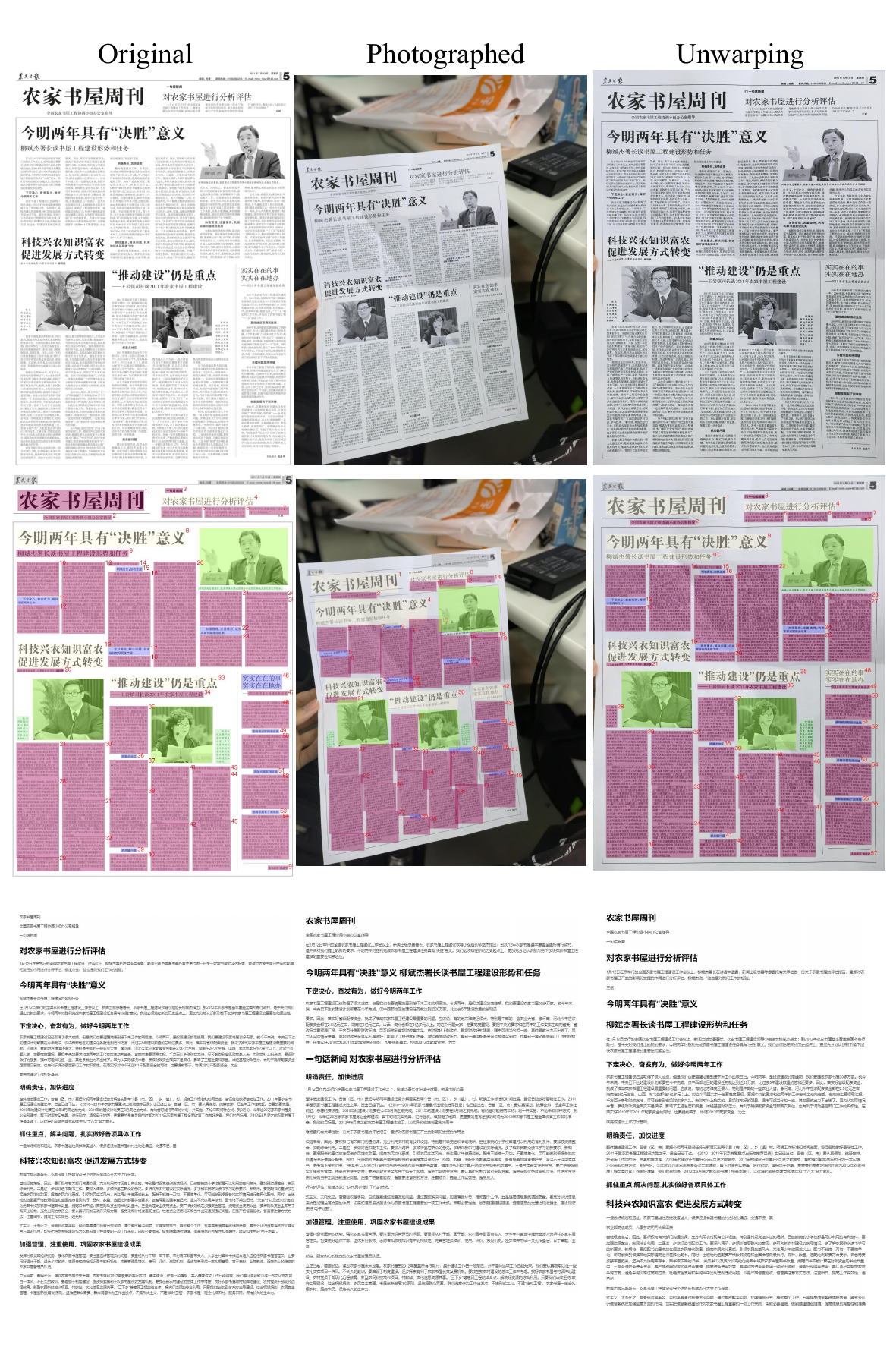}
    \caption{Qualitative visualization of PaddleOCR-VL's output on the Original document versus the Photographed and Unwarped versions.}
  \label{fig:parsing_visualization_1_paddle}
\end{figure*}

\begin{figure*}[!t]
  \centering
  \includegraphics[width=0.88\textwidth, clip, trim={0 3mm 0 6mm}]{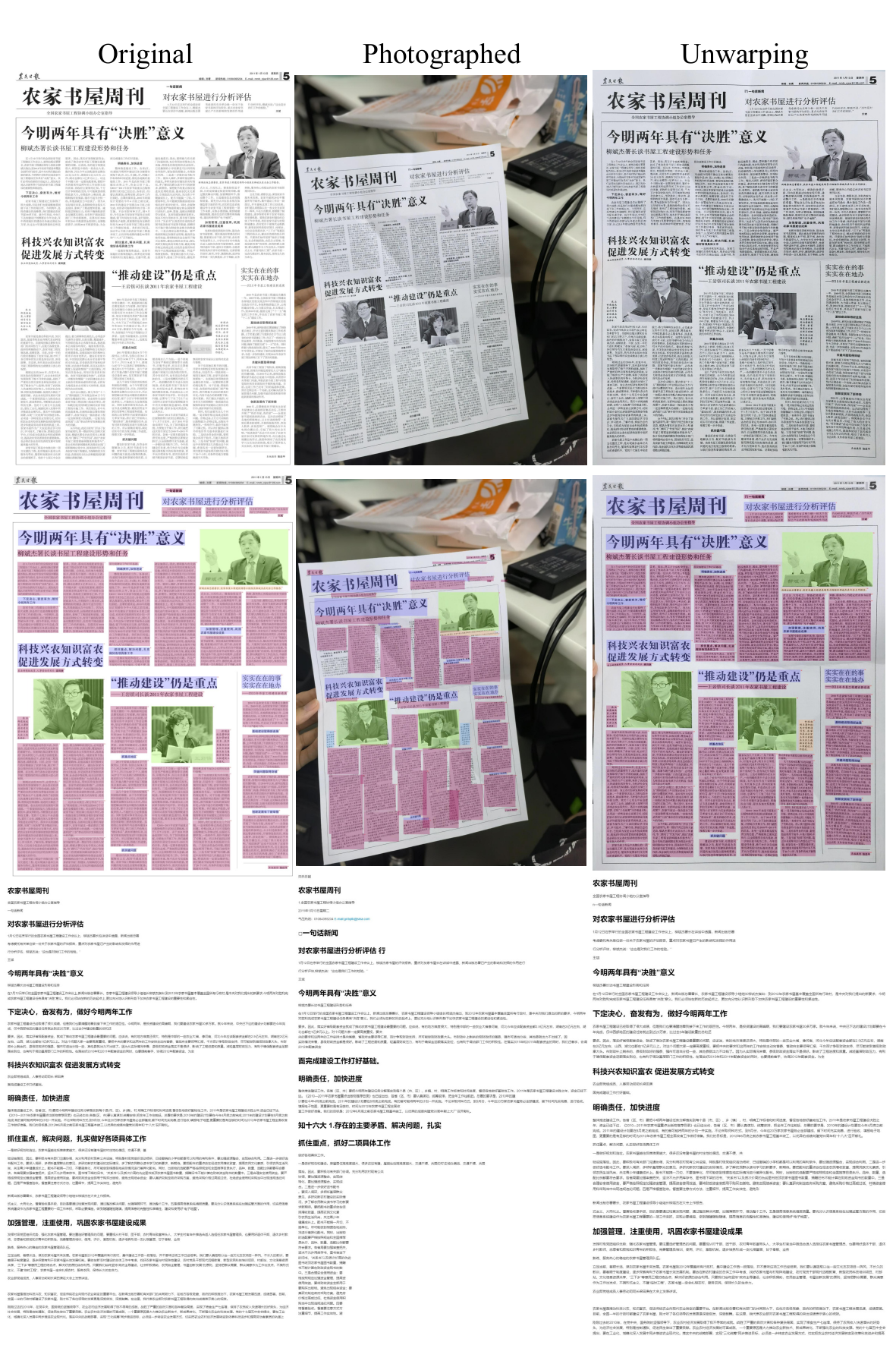}
    \caption{Qualitative visualization of MinerU2.5's output on the Original document versus the Photographed and Unwarped versions.}
  \label{fig:parsing_visualization_1_mineru}
\end{figure*}

\begin{figure*}[!t]
  \centering
  \includegraphics[width=0.88\textwidth, clip, trim={0 3mm 0 6mm}]{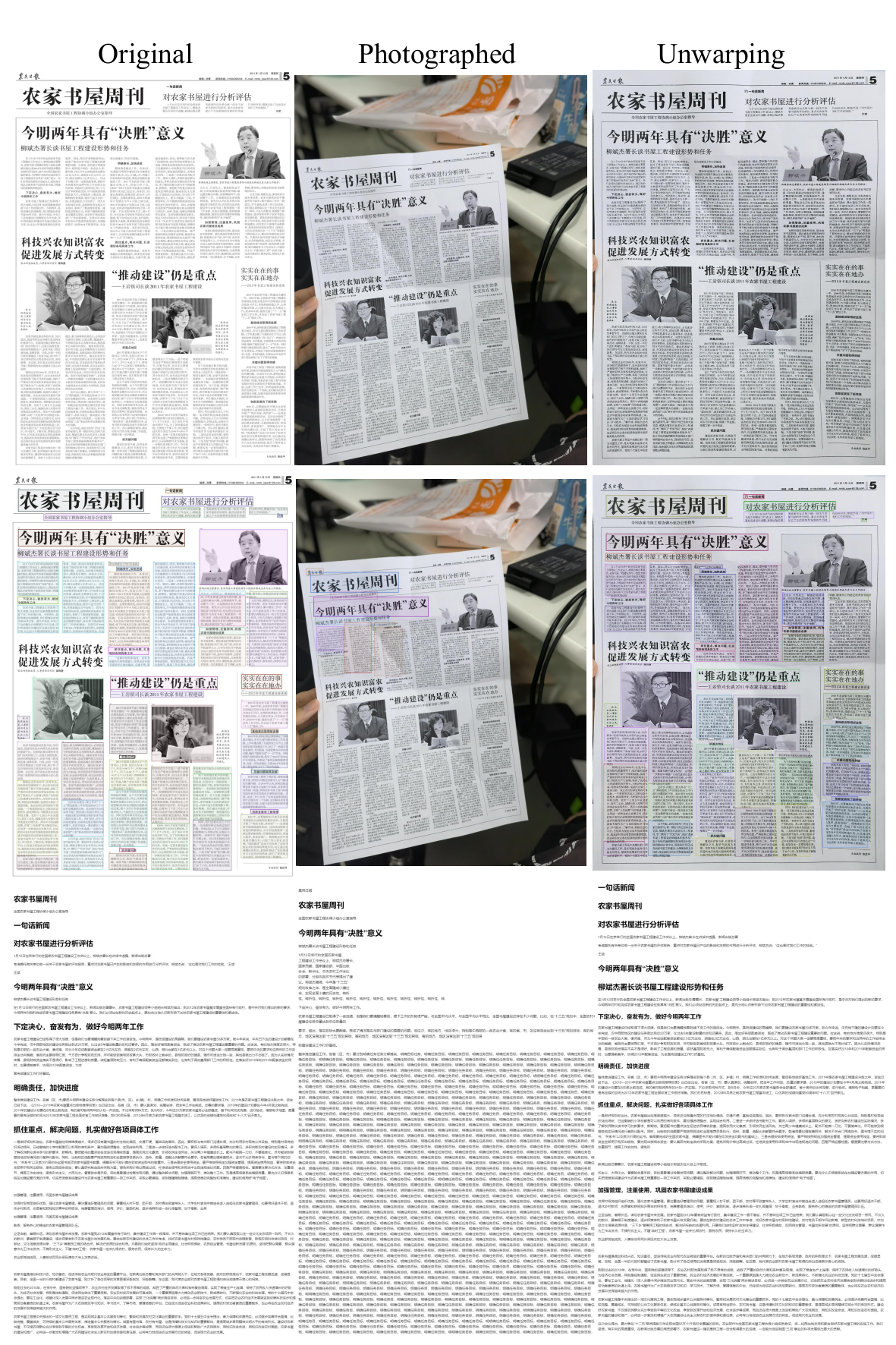}
    \caption{Qualitative visualization of DeepSeek-OCR's output on the Original document versus the Photographed and Unwarped versions.}
  \label{fig:parsing_visualization_1_deepseekocr}
\end{figure*}

\begin{figure*}[!t]
  \centering
  \includegraphics[width=0.88\textwidth, clip, trim={0 3mm 0 6mm}]{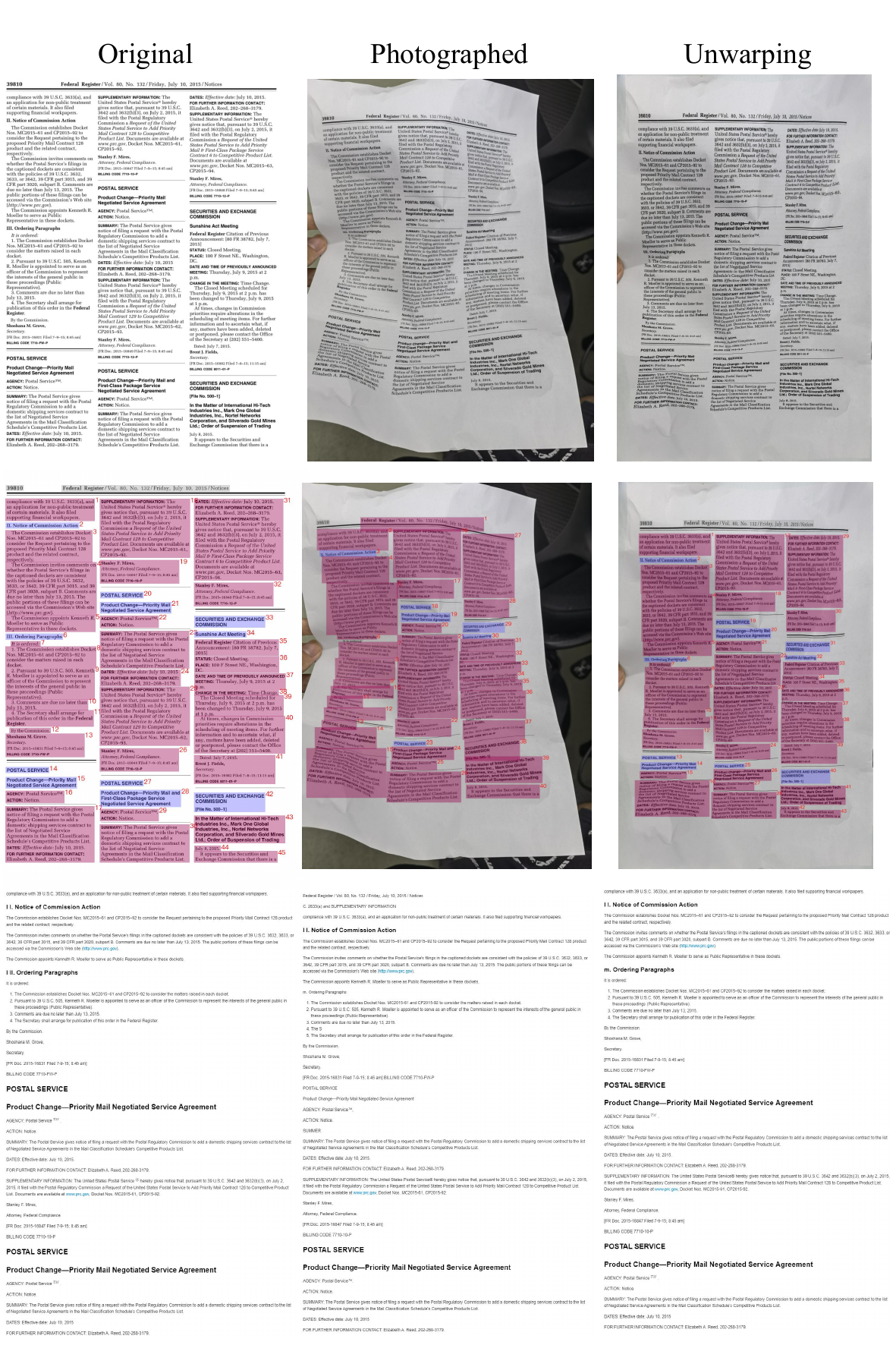}
    \caption{Qualitative visualization of PaddleOCR-VL's output on the Original document versus the Photographed and Unwarped versions.}
  \label{fig:parsing_visualization_2_paddle}
\end{figure*}

\begin{figure*}[!t]
  \centering
  \includegraphics[width=0.88\textwidth, clip, trim={0 3mm 0 6mm}]{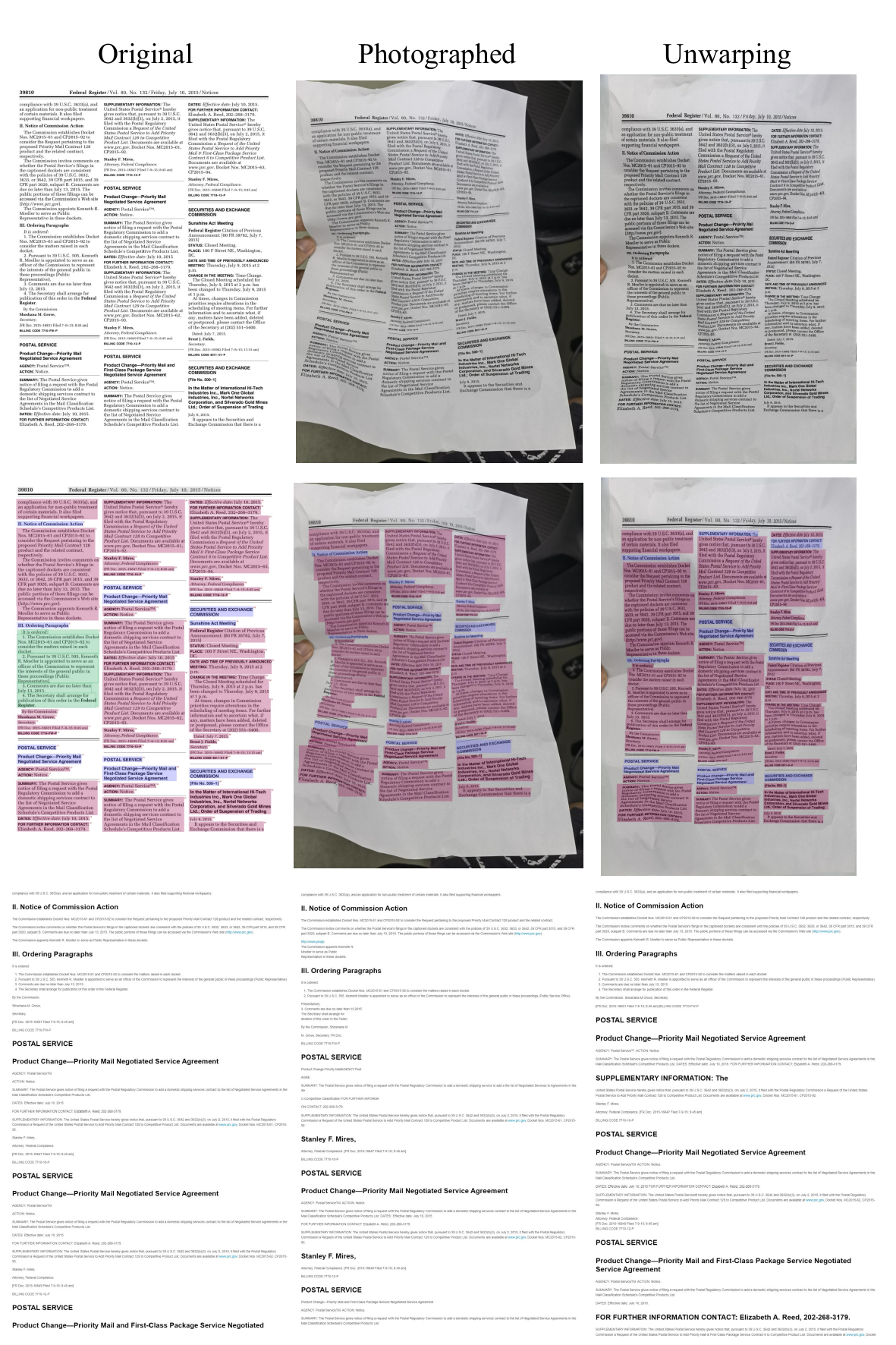}
    \caption{Qualitative visualization of MinerU2.5's output on the Original document versus the Photographed and Unwarped versions.}
  \label{fig:parsing_visualization_2_mineru}
\end{figure*}

\begin{figure*}[!t]
  \centering
  \includegraphics[width=0.88\textwidth, clip, trim={0 3mm 0 6mm}]{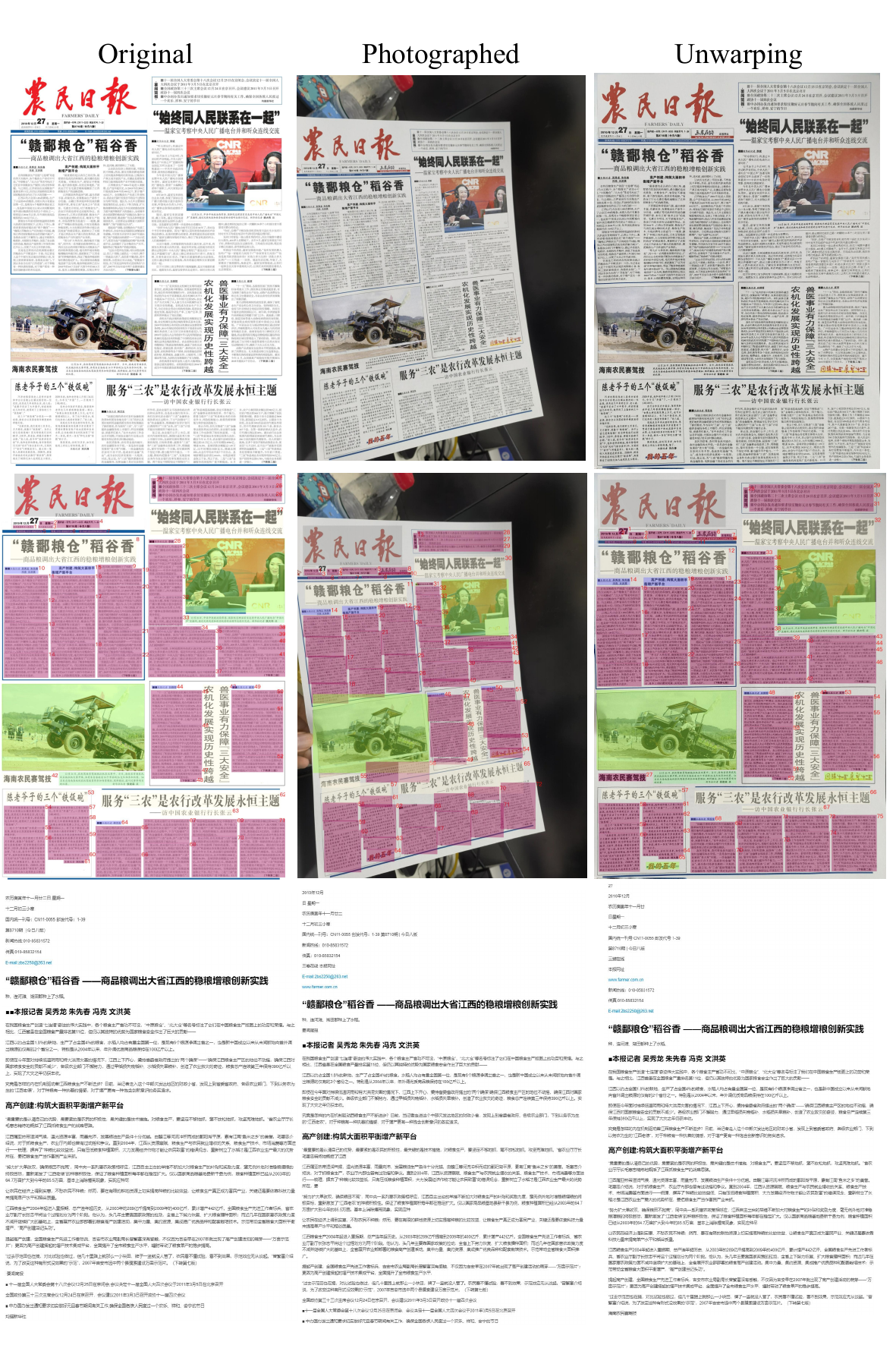}
    \caption{Qualitative visualization of PaddleOCR-VL's output on the Original document versus the Photographed and Unwarped versions.}
  \label{fig:parsing_visualization_3_paddle}
\end{figure*}

\begin{figure*}[!t]
  \centering
  \includegraphics[width=0.88\textwidth, clip, trim={0 3mm 0 6mm}]{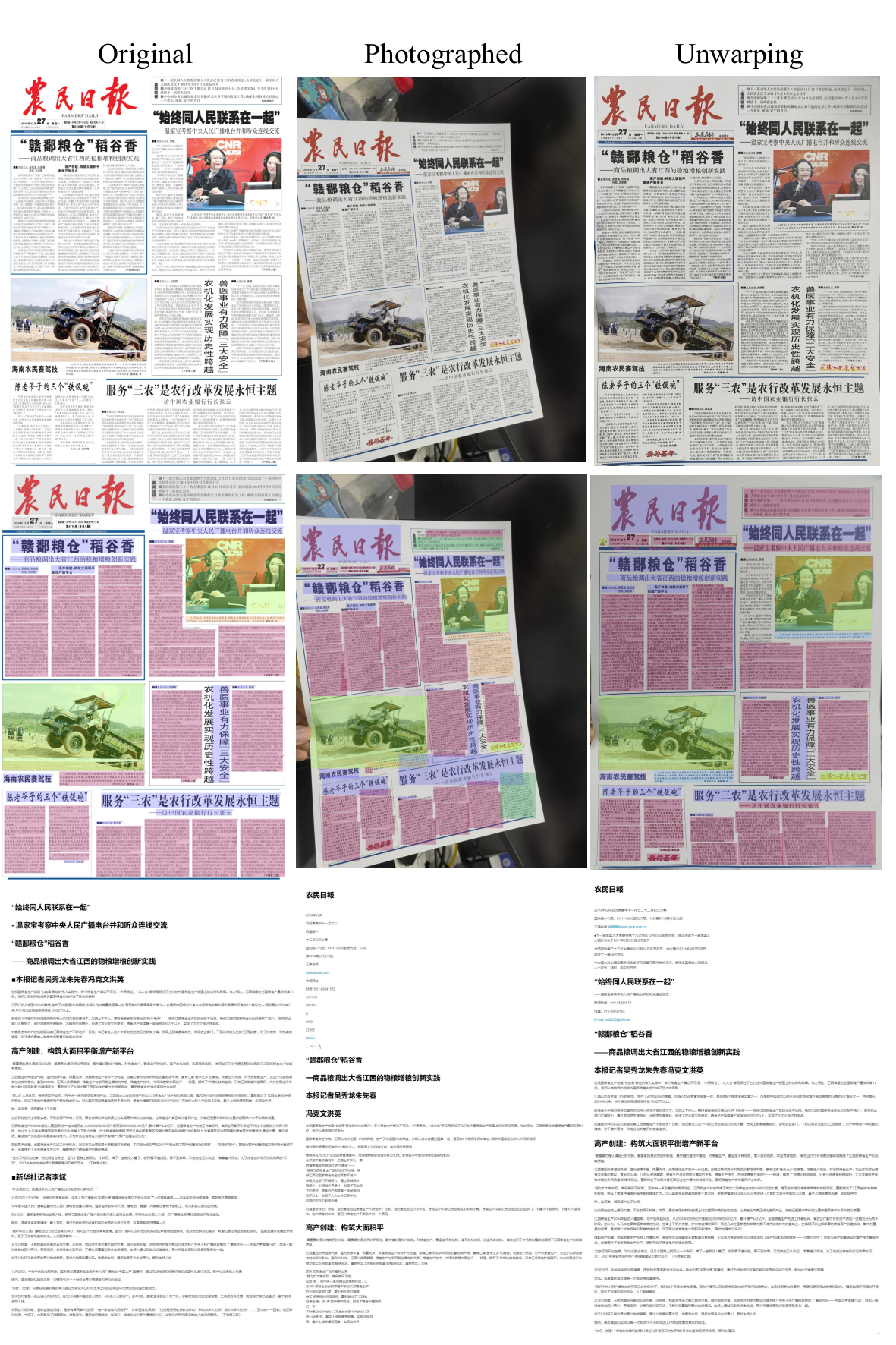}
    \caption{Qualitative visualization of MinerU2.5's output on the Original document versus the Photographed and Unwarped versions.}
  \label{fig:parsing_visualization_3_mineru}
\end{figure*}

\begin{figure*}[!t]
  \centering
  \includegraphics[width=0.88\textwidth, clip, trim={0 3mm 0 6mm}]{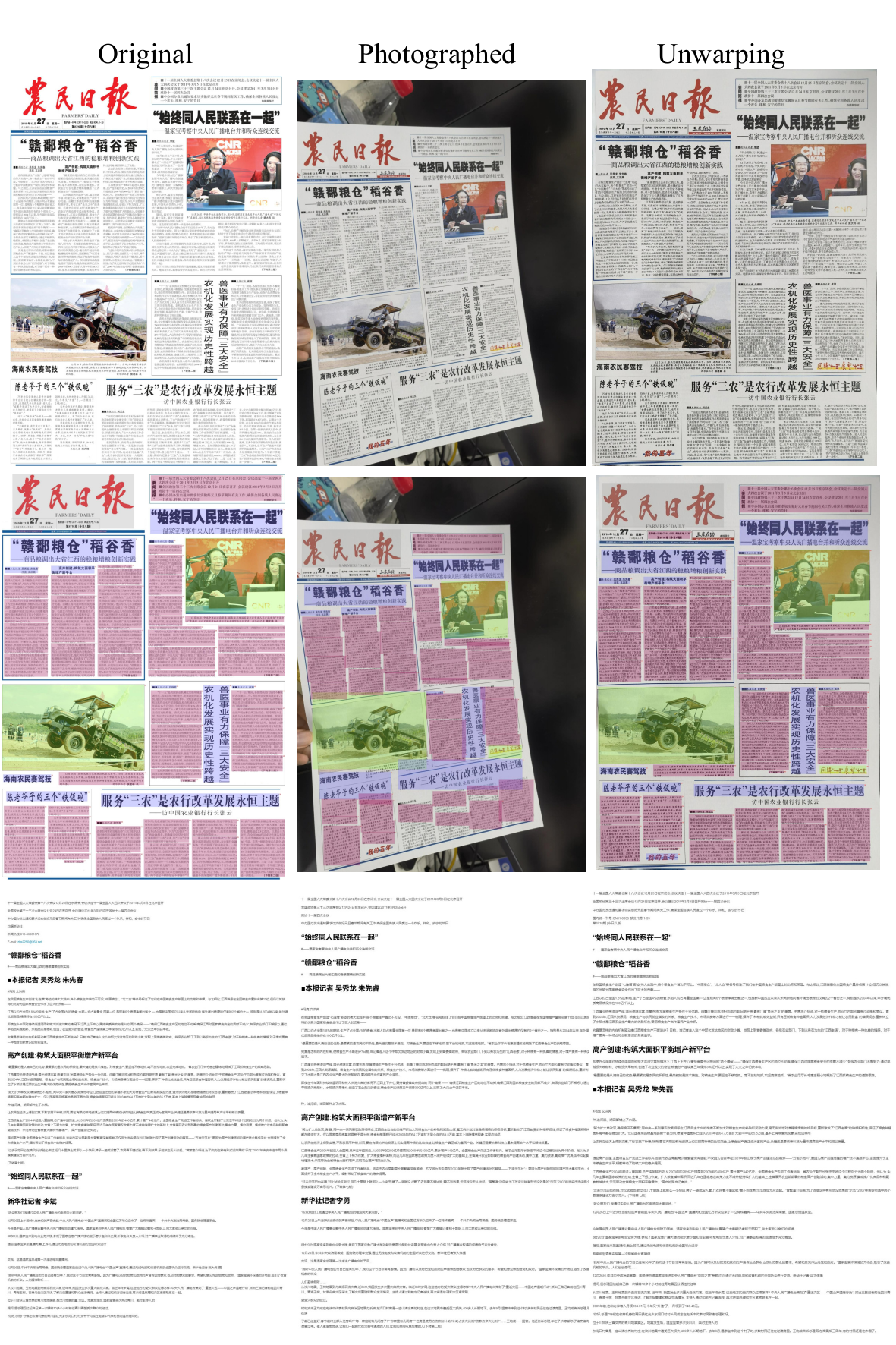}
    \caption{Qualitative visualization of MonkeyOCR's output on the Original document versus the Photographed and Unwarped versions.}
  \label{fig:parsing_visualization_3_monkeyocr}
\end{figure*}

\begin{figure*}[!t]
  \centering
  \includegraphics[width=0.88\textwidth, clip, trim={0 3mm 0 6mm}]{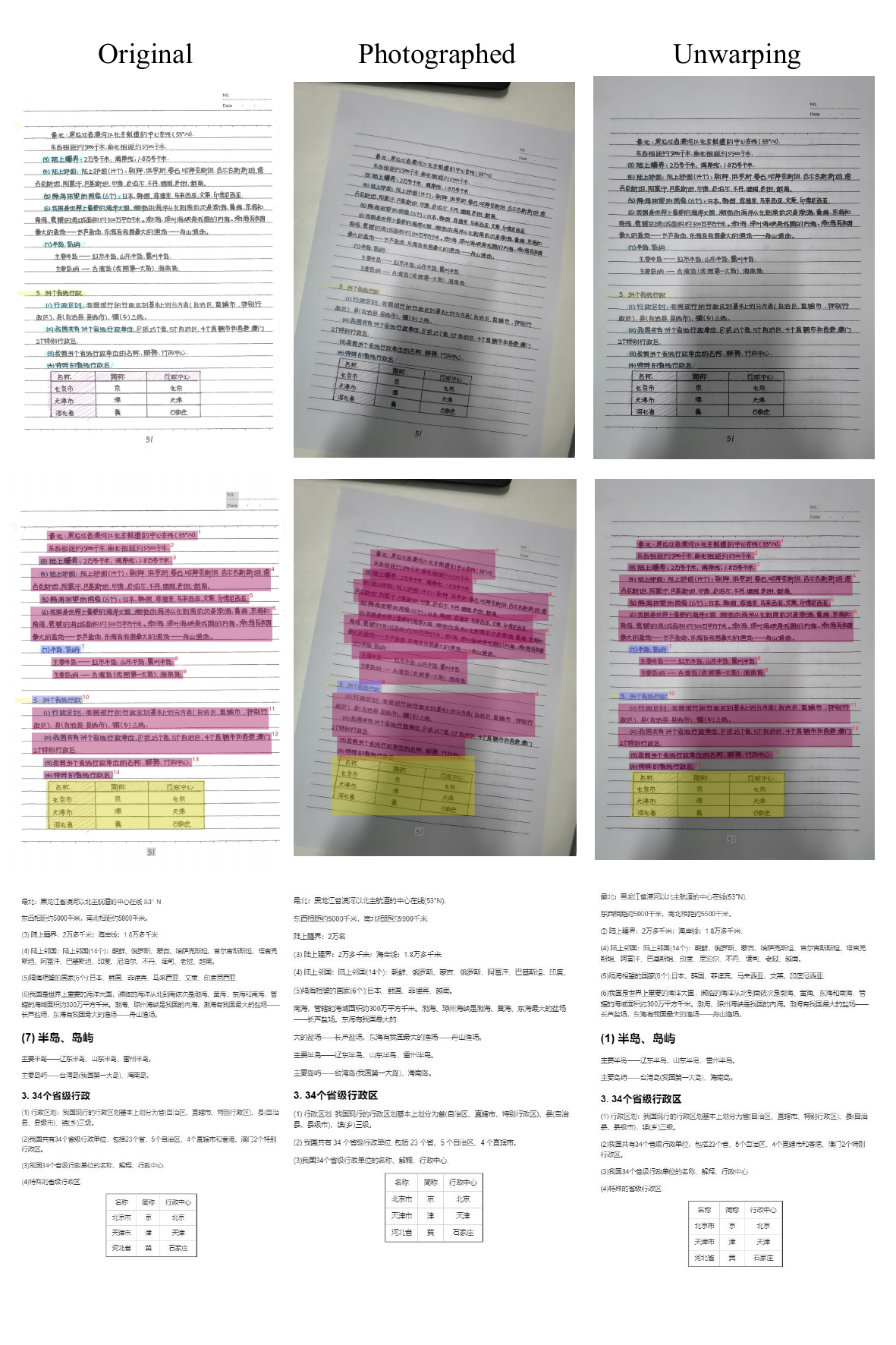}
    \caption{Qualitative visualization of PaddleOCR-VL's output on the Original document versus the Photographed and Unwarped versions.}
  \label{fig:parsing_visualization_4_paddle}
\end{figure*}

\begin{figure*}[!t]
  \centering
  \includegraphics[width=0.88\textwidth, clip, trim={0 3mm 0 6mm}]{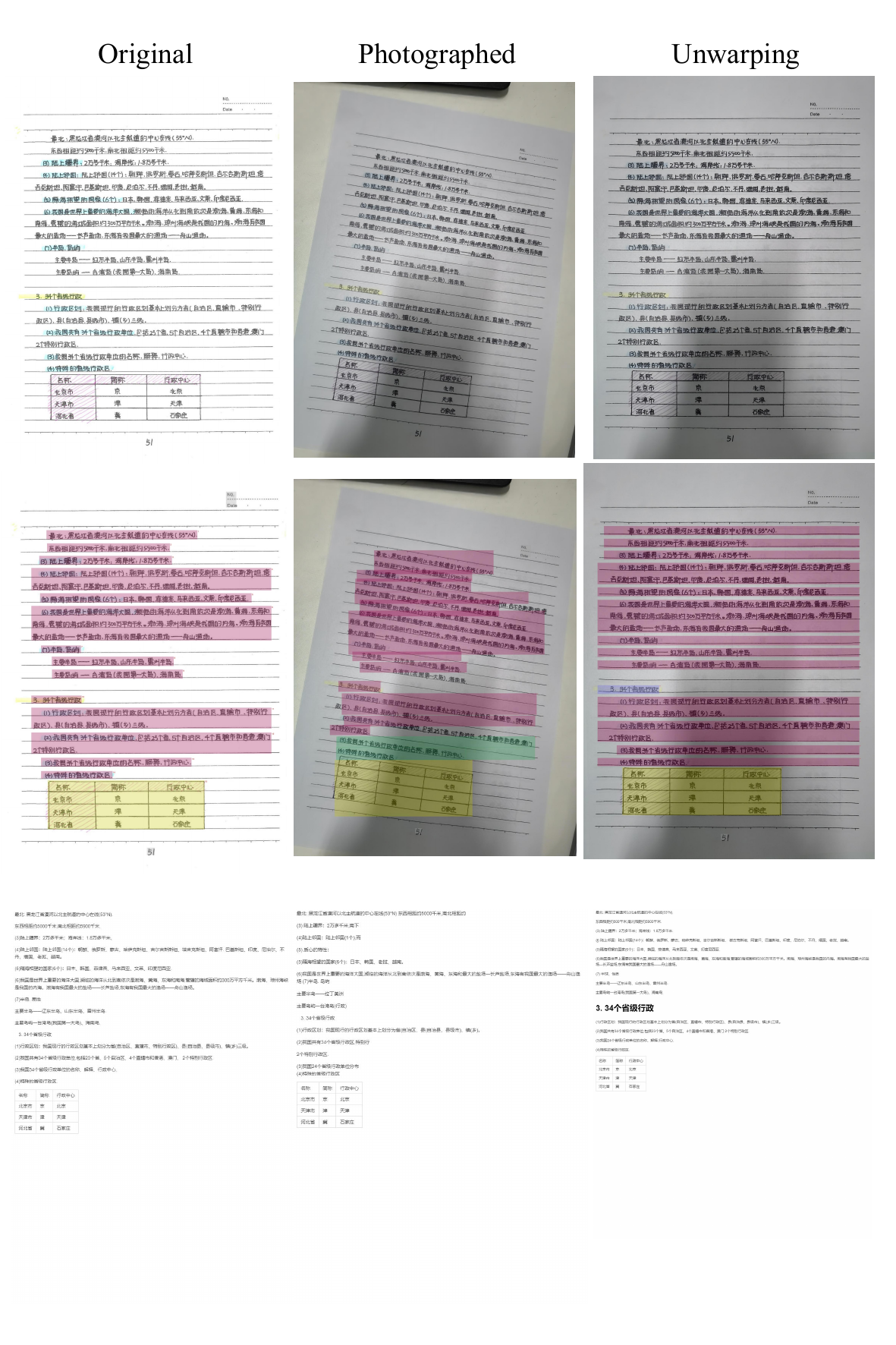}
    \caption{Qualitative visualization of MinerU2.5's output on the Original document versus the Photographed and Unwarped versions.}
  \label{fig:parsing_visualization_4_mineru}
\end{figure*}

\begin{figure*}[!t]
  \centering
  \includegraphics[width=0.88\textwidth, clip, trim={0 3mm 0 6mm}]{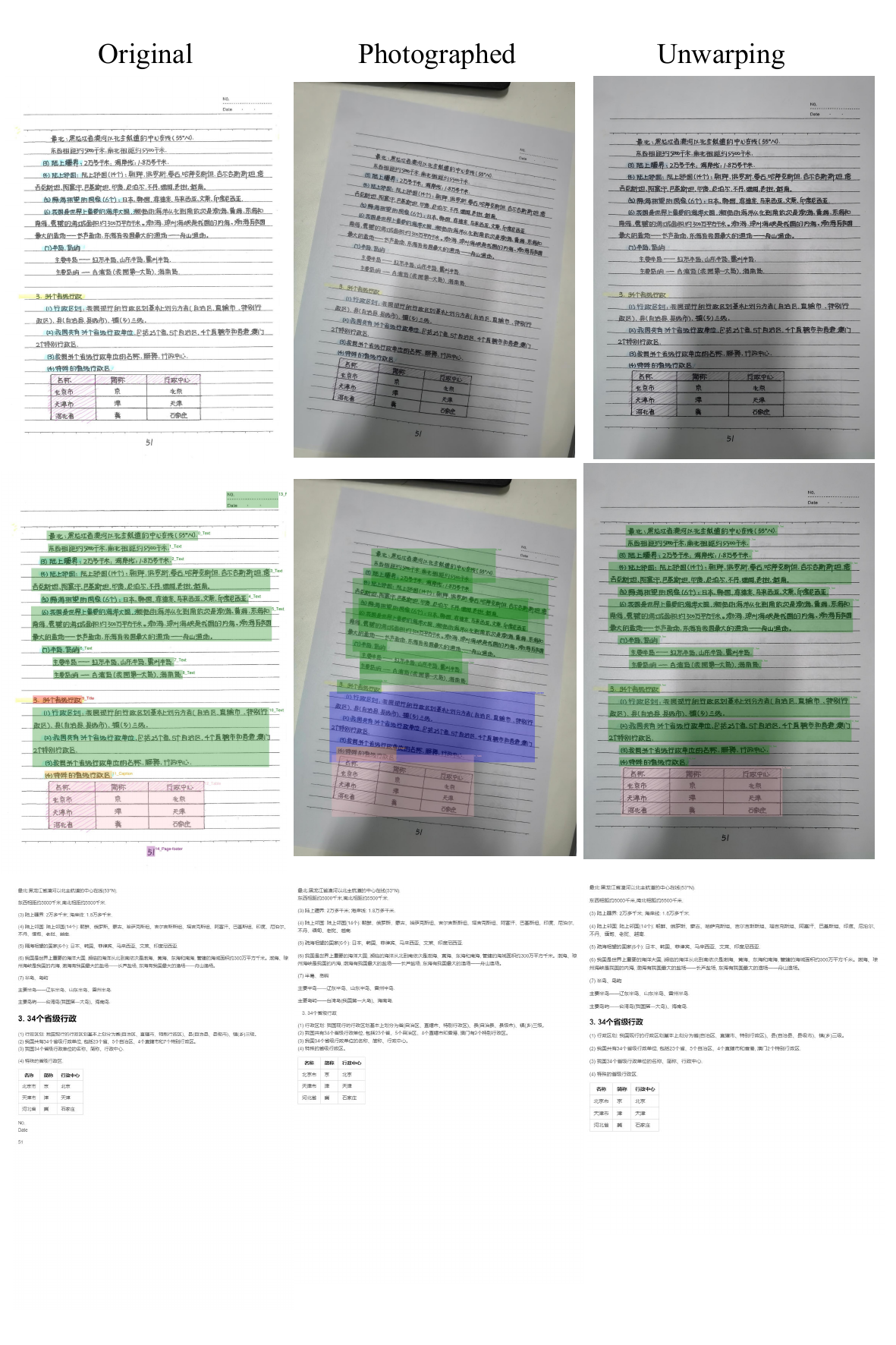}
    \caption{Qualitative visualization of dots.ocr's output on the Original document versus the Photographed and Unwarped versions.}
  \label{fig:parsing_visualization_4_dotsocr}
\end{figure*}

\begin{figure*}[!t]
  \centering
  \includegraphics[width=0.88\textwidth, clip, trim={0 3mm 0 6mm}]{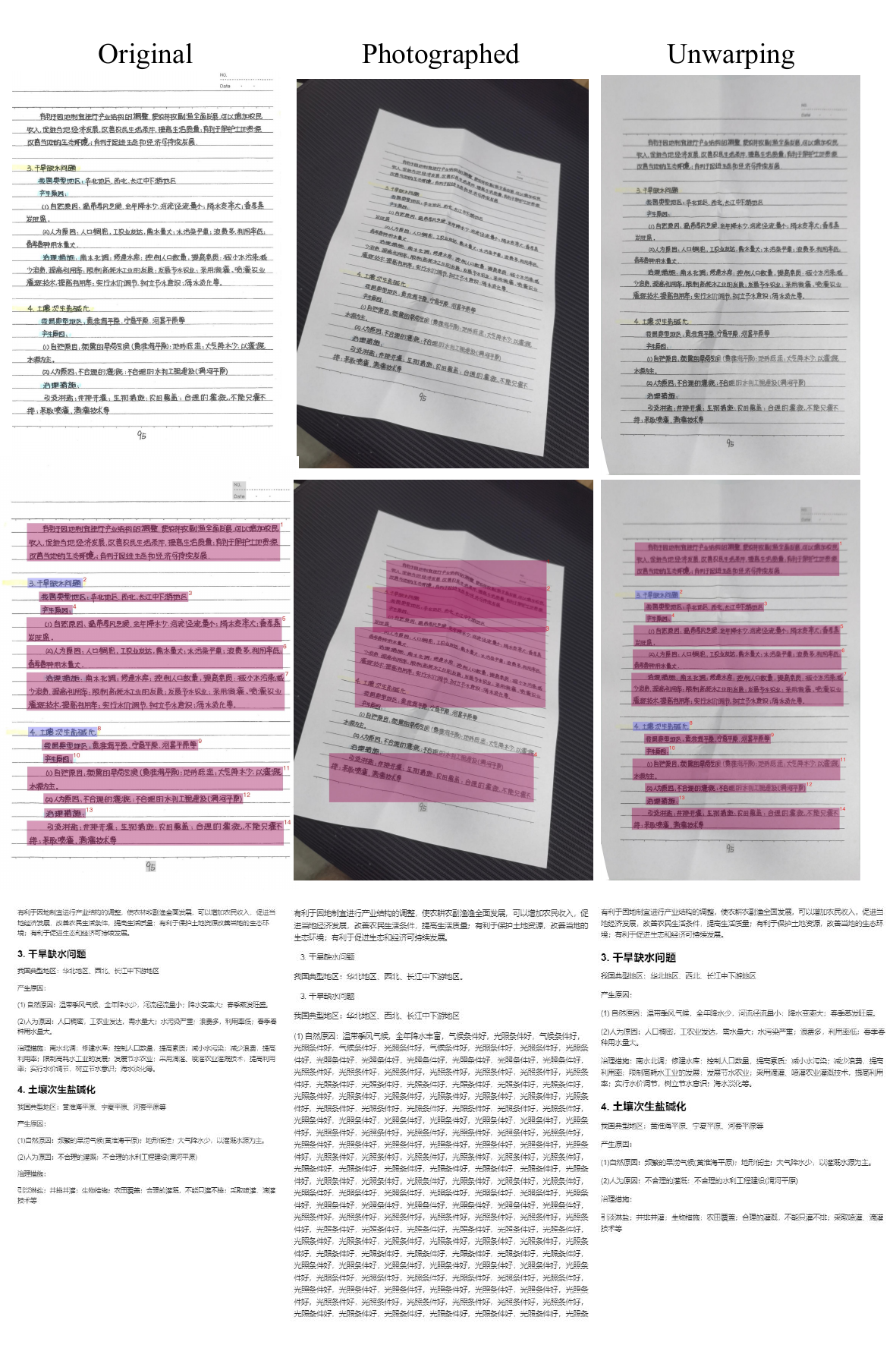}
    \caption{Qualitative visualization of PaddleOCR-VL's output on the Original document versus the Photographed and Unwarped versions.}
  \label{fig:parsing_visualization_5_paddle}
\end{figure*}

\begin{figure*}[!t]
  \centering
  \includegraphics[width=0.88\textwidth, clip, trim={0 3mm 0 6mm}]{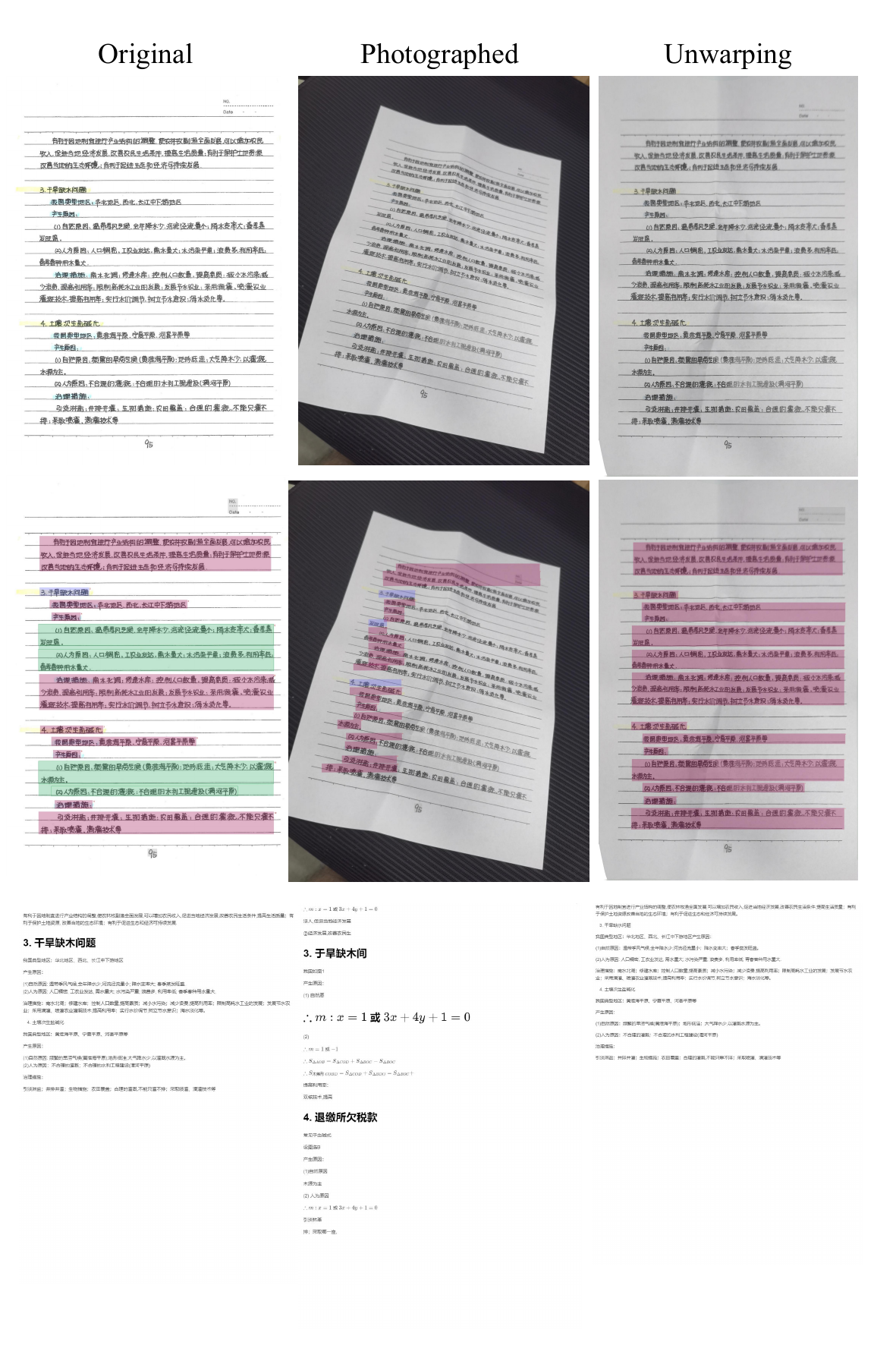}
    \caption{Qualitative visualization of MinerU2.5's output on the Original document versus the Photographed and Unwarped versions.}
  \label{fig:parsing_visualization_5_mineru}
\end{figure*}

The specific prompts utilized in our experiments are illustrated in Fig.~\ref{fig:prompt}. We extend our evaluation to En-De/Fr/Ru and Zh-De/Fr/Ru language pairs. Detailed metrics are provided in Tab.~\ref{tab:translation_result_de}, ~\ref{tab:translation_result_fr}, and ~\ref{tab:translation_result_ru}.

\subsection{Modality Gap Across Languages}
A consistent modality gap is observed across all language pairs. Comparing the \textit{Text} baseline with \textit{Original-Simple}, performance drops universally. For instance, in the En-Fr translation task (Tab.~\ref{tab:translation_result_fr}), Qwen-VL-Max's~\cite{Qwen-VL} BLEU score drops from 68.65 to 46.02. In the En-Ru translation task (Tab.~\ref{tab:translation_result_ru}), InternVL3-2B~\cite{zhu2025internvl3} fails almost completely, with its BLEU score dropping from 37.09 to 8.15, indicating that low-resource scripts exacerbate more wchallenges.

\subsection{Efficacy of Chain-of-Thought Prompting}
Our results demonstrate that the CoT strategy, which explicitly instructs the model to perform OCR before translation, is a robust mechanism for mitigating the modality gap. In Tab.~\ref{tab:translation_result_fr}, Qwen3-VL-4B~\cite{qwen3vl} improves its En-Fr BLEU score by +5.47 points using CoT. Similarly, in Tab.~\ref{tab:translation_result_de}, InternVL3.5-2B~\cite{wang2025internvl3} sees a +7.03 BLEU gain on En-De. 

\subsection{Qualitative Visualizations}
\begin{figure*}[!t]
  \centering
  \includegraphics[width=0.9\textwidth, clip, trim={0 8mm 0 6mm}]{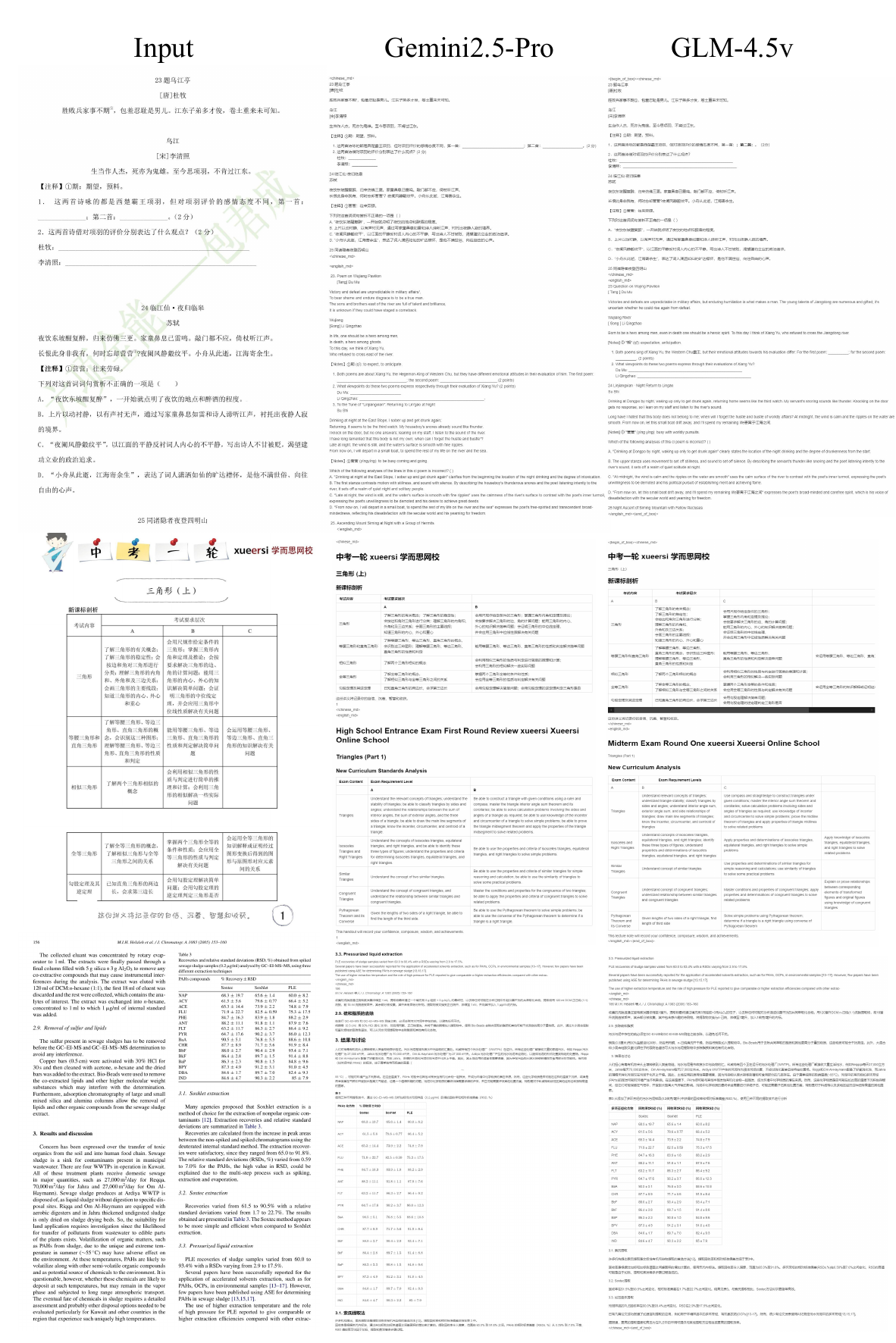}
    \caption{Qualitative comparison of Gemini 2.5-Pro and GLM-4.5v on Zh-En and En-Zh document translation tasks.}
  \label{fig:translation_visualization_1}
\end{figure*}

\begin{figure*}[!t]
  \centering
  \includegraphics[width=0.9\textwidth, clip, trim={0 8mm 0 6mm}]{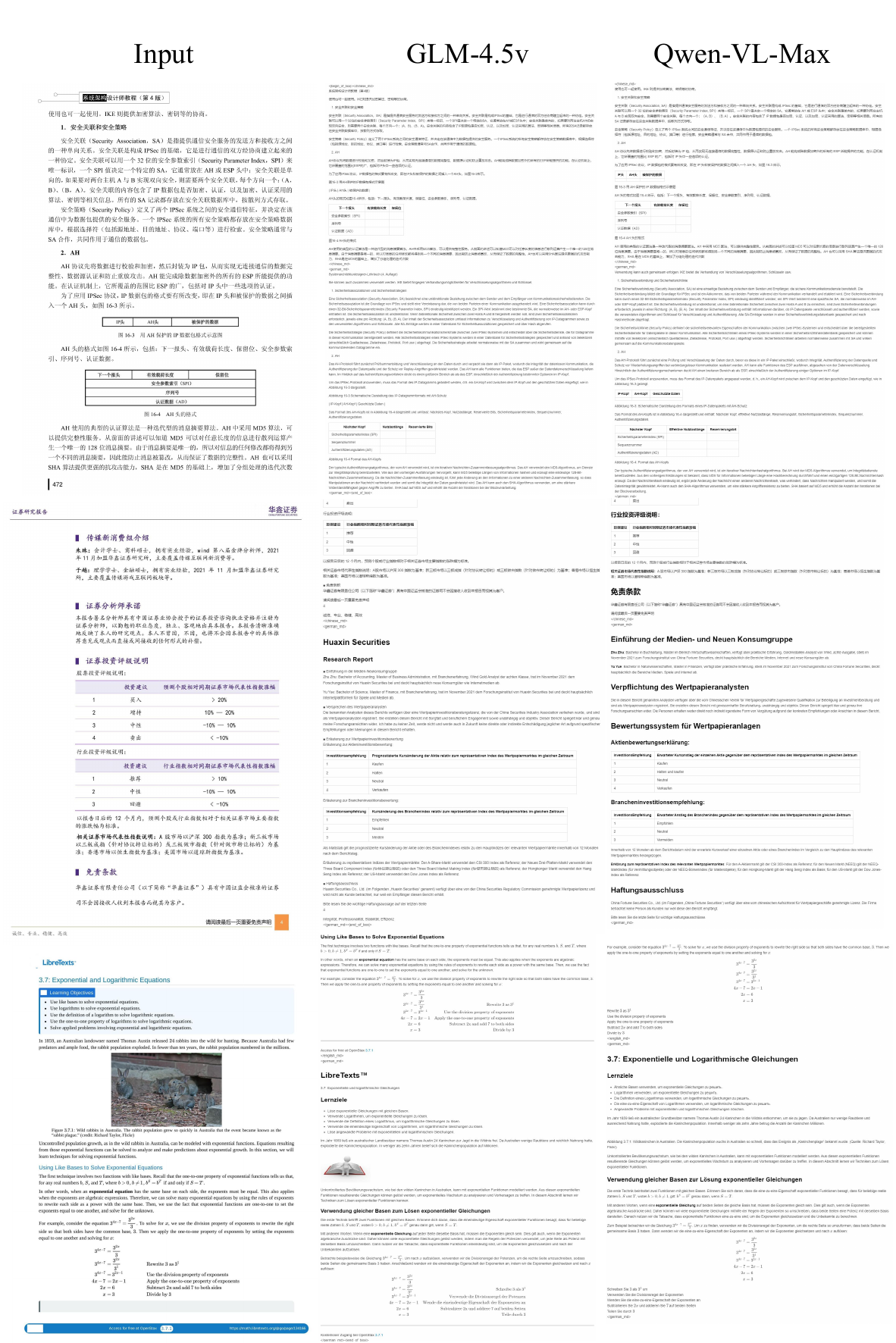}
    \caption{Qualitative comparison of GLM-4.5v and Qwen-VL-Max on Zh-De and En-De translation tasks.}
  \label{fig:translation_visualization_2}
\end{figure*}

\begin{figure*}[!t]
  \centering
  \includegraphics[width=0.9\textwidth, clip, trim={0 8mm 0 6mm}]{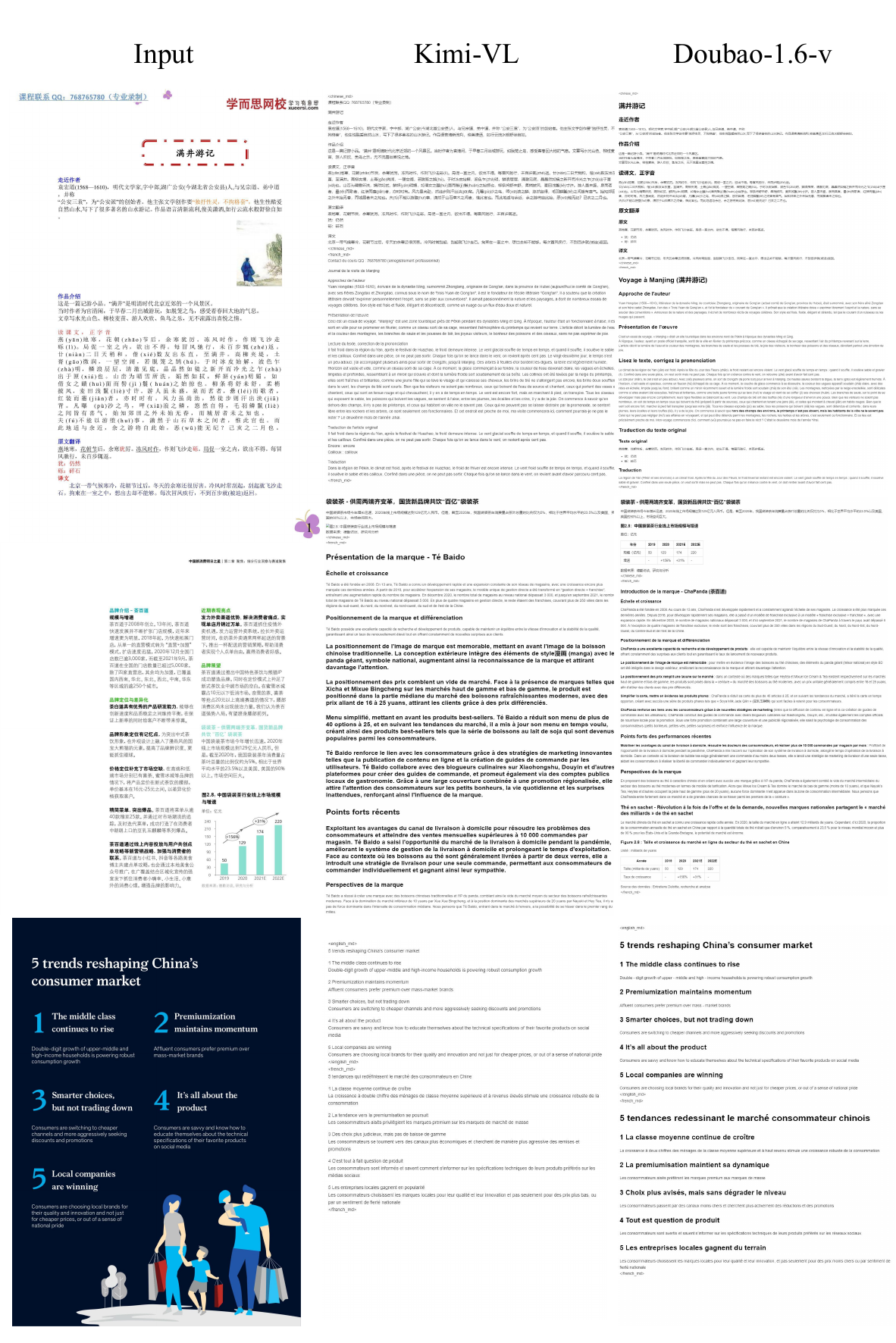}
    \caption{Qualitative comparison of Kimi-VL and Doubao-1.6-v on Zh-Fr and En-Fr translation tasks.}
  \label{fig:translation_visualization_3}
\end{figure*}

\begin{figure*}[!t]
  \centering
  \includegraphics[width=0.9\textwidth, clip, trim={0 10mm 0 6mm}]{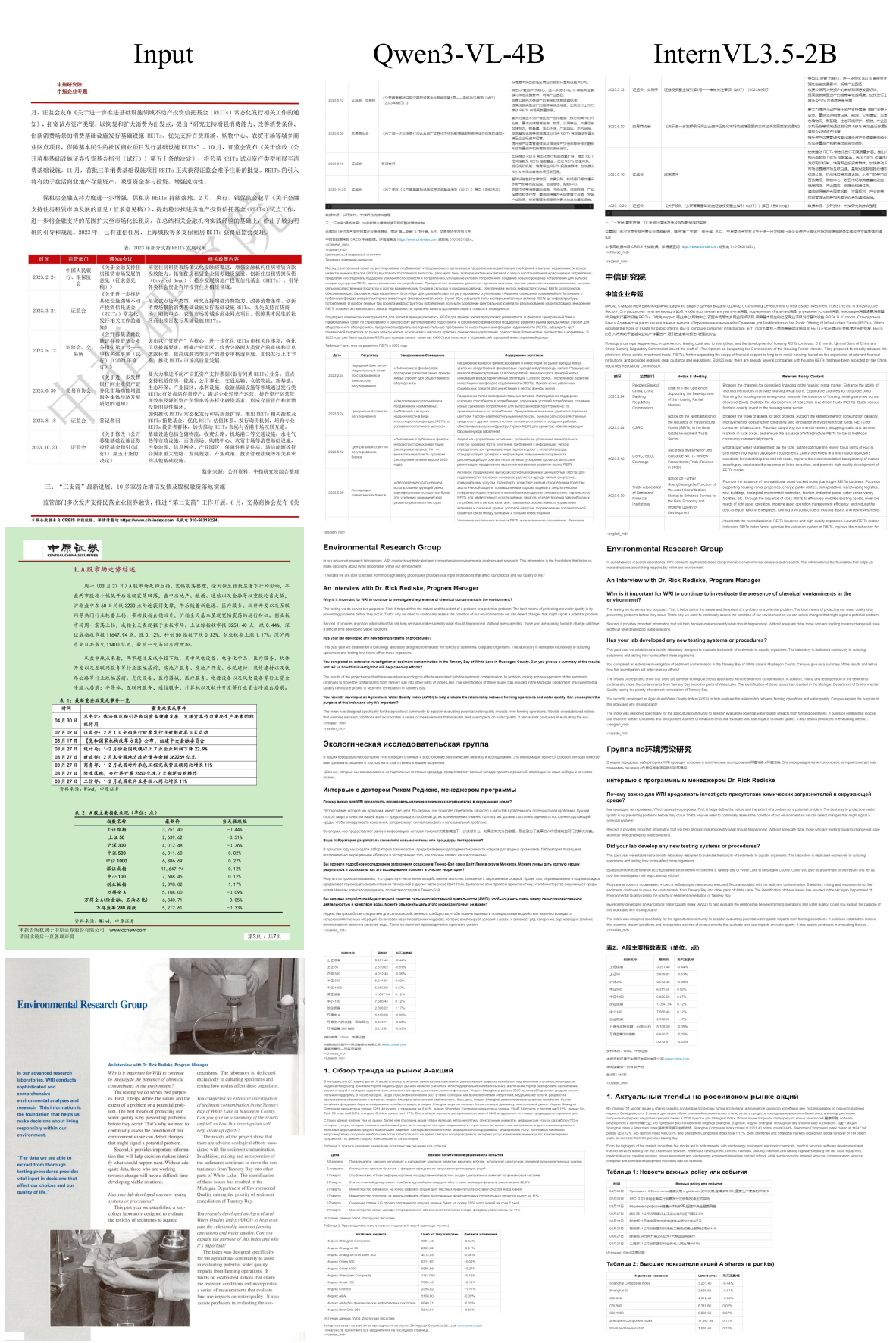}
    \caption{Qualitative comparison of Qwen3-VL-4B and InternVL3.5-2B on Zh-Ru and En-Ru translation tasks.}
  \label{fig:translation_visualization_4}
\end{figure*}

\begin{figure*}[!t]
  \centering
  \includegraphics[width=0.9\textwidth, clip, trim={0 15mm 0 6mm}]{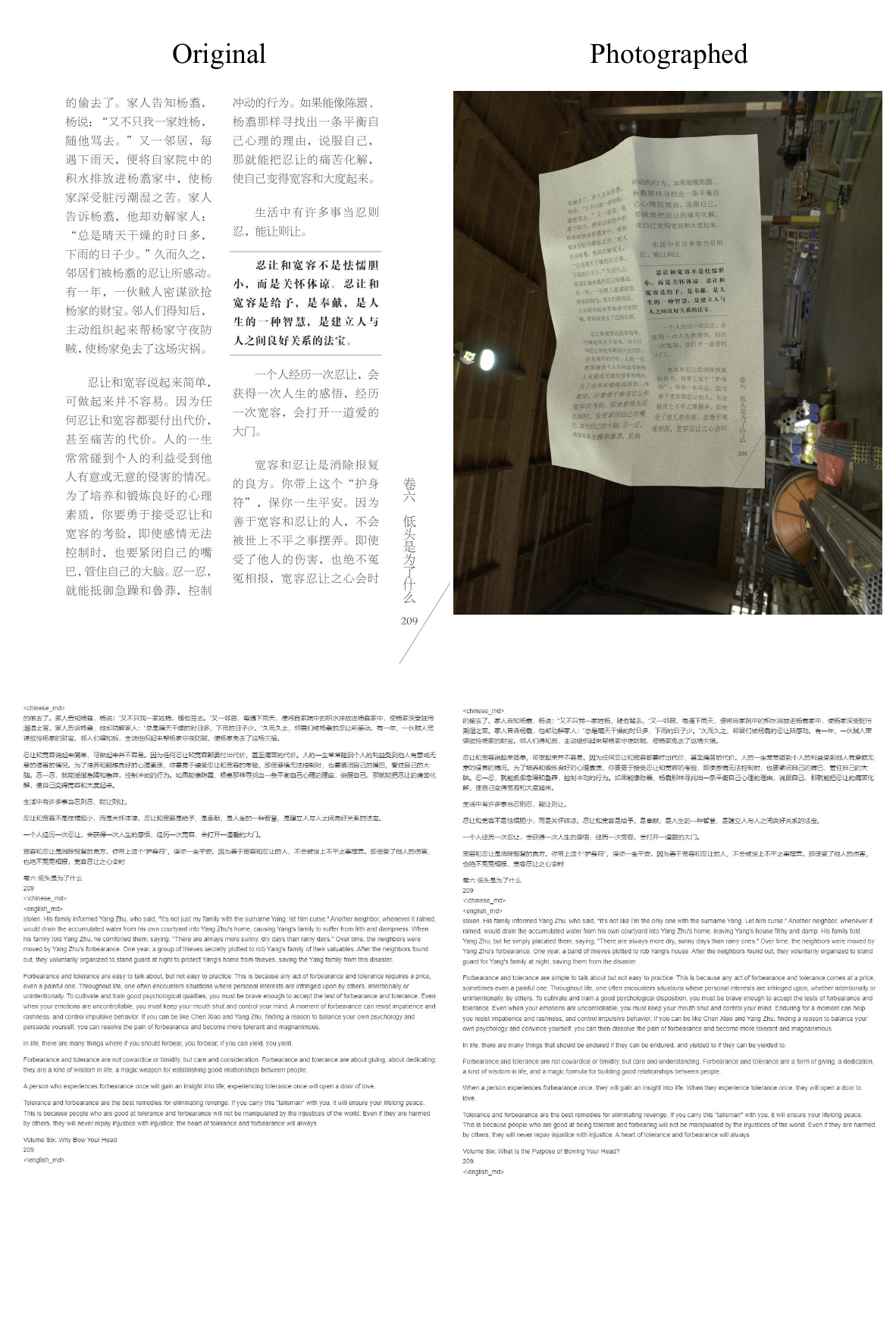}
    \caption{Impact of photographic distortion on Gemini 2.5-Pro's translation performance for the Zh-En task.}
  \label{fig:translation_visualization_5}
\end{figure*}

\begin{figure*}[!t]
  \centering
  \includegraphics[width=0.9\textwidth, clip, trim={0 15mm 0 6mm}]{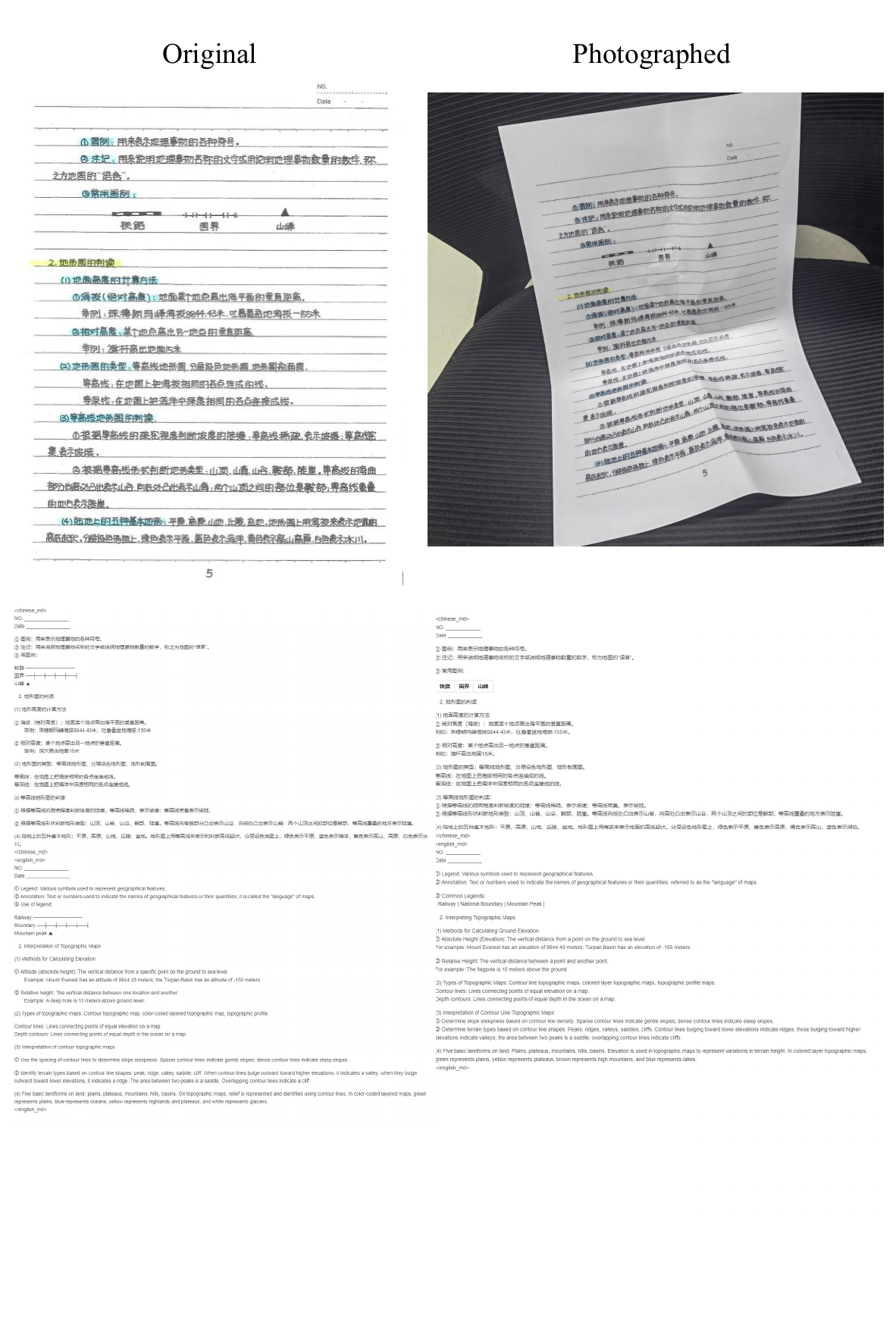}
    \caption{Impact of photographic distortion on Doubao-1.6-v's translation performance for the Zh-En task.}
  \label{fig:translation_visualization_6}
\end{figure*}

\begin{figure*}[!t]
  \centering
  \includegraphics[width=0.9\textwidth, clip, trim={0 15mm 0 6mm}]{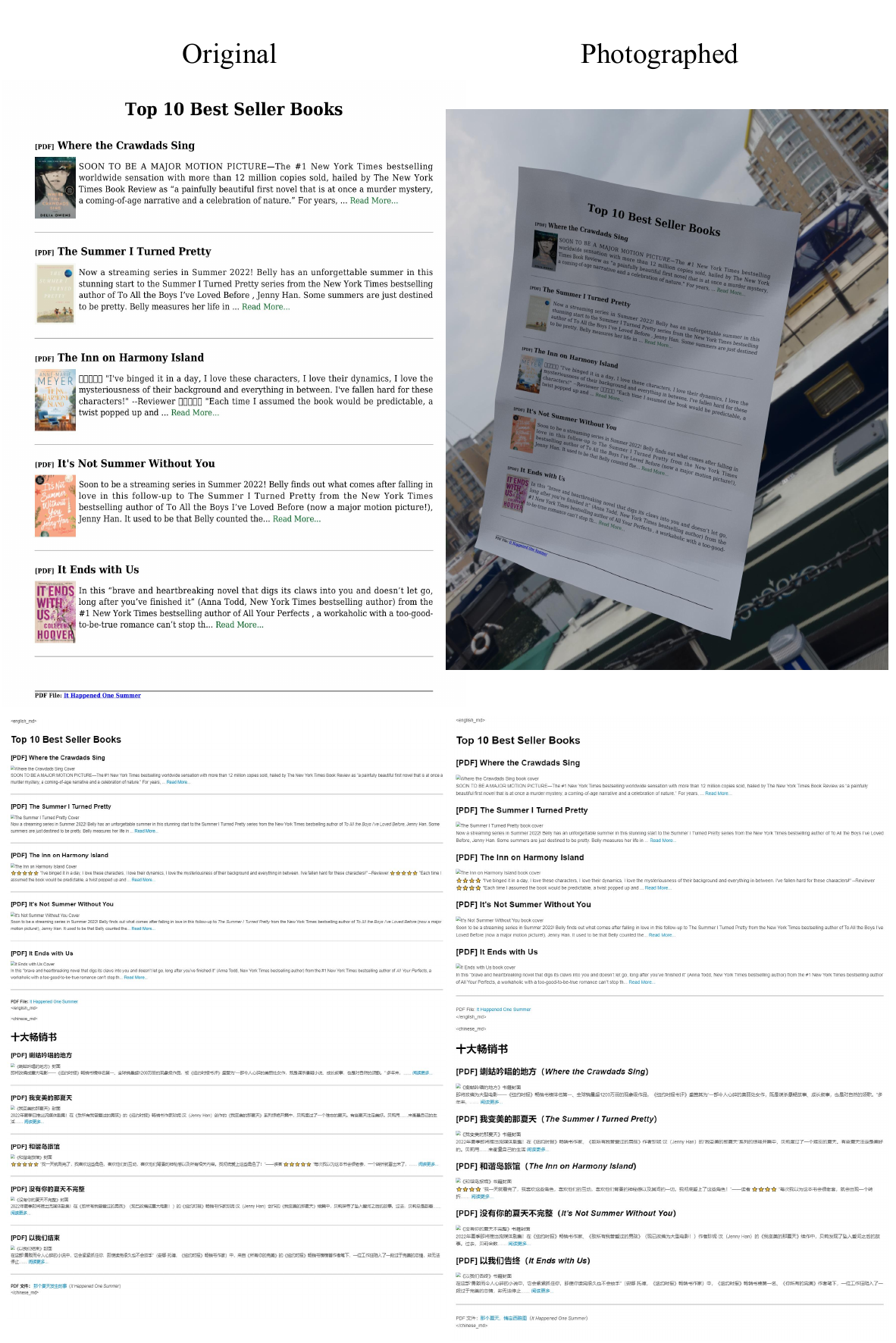}
    \caption{Impact of photographic distortion on GLM-4.5v's translation performance for the En-Zh task.}
  \label{fig:translation_visualization_7}
\end{figure*}

\begin{figure*}[!t]
  \centering
  \includegraphics[width=0.9\textwidth, clip, trim={0 15mm 0 6mm}]{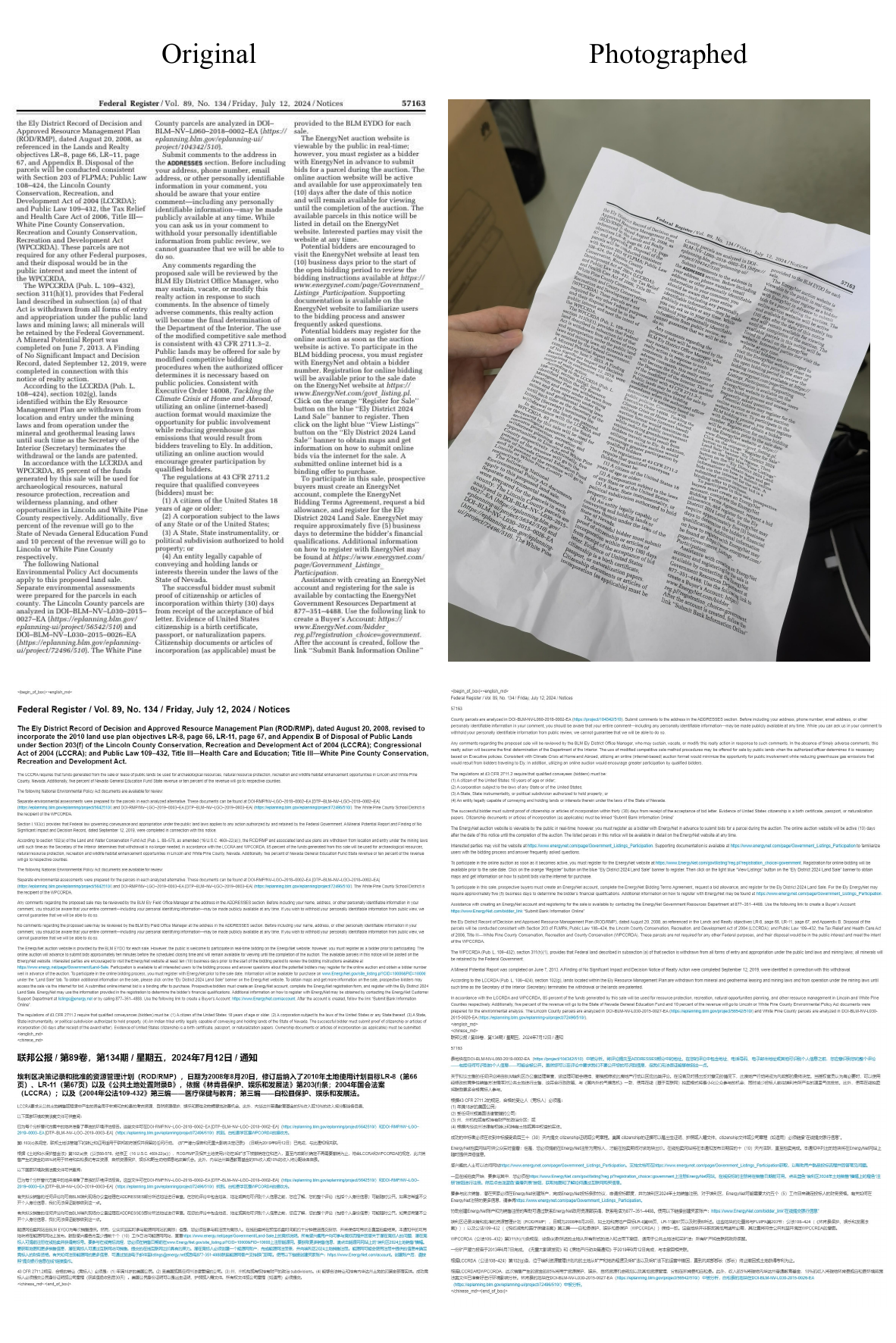}
    \caption{Impact of photographic distortion on Qwen-VL-Max's translation performance for the En-Zh task.}
  \label{fig:translation_visualization_8}
\end{figure*}

Fig.~\ref{fig:translation_visualization_1} to Fig.~\ref{fig:translation_visualization_4} showcase the translation outputs across various language pairs, demonstrating the models' ability to handle diverse scripts and complex layouts. Furthermore, Fig.~\ref{fig:translation_visualization_5} to Fig.~\ref{fig:translation_visualization_8} explicitly contrast model performance on digital-born documents versus photographed images. While models often maintain semantic accuracy on clean inputs, photographic artifacts can lead to hallucinations, omitted text, or layout misalignment, reinforcing the quantitative performance drops observed in our benchmarks.

\begin{table*}[!htbp]
\centering
\captionsetup{font=small} 

\scriptsize 
\setlength{\tabcolsep}{1pt} 
\renewcommand{\arraystretch}{1.00} 

\begin{tabular*}{\textwidth}{@{\extracolsep{\fill}} l l l ll ll ll ll ll ll @{}}
\toprule
\multirow{2}{*}{\textbf{Type}} & \multirow{2}{*}{\textbf{Model}} & \multirow{2}{*}{\textbf{Scene}} & \multicolumn{2}{c}{\textbf{Overall$^{\mathbf{Edit}\downarrow}$}} & \multicolumn{2}{c}{\textbf{Text$^{\mathbf{Edit}\downarrow}$}} & \multicolumn{2}{c}{\textbf{Formula$^{\mathbf{Edit}\downarrow}$}} & \multicolumn{2}{c}{\textbf{Table$^{\mathbf{TEDS}\uparrow}$}} & \multicolumn{2}{c}{\textbf{Table$^{\mathbf{Edit}\downarrow}$}} & \multicolumn{2}{c}{\textbf{Read Order$^{\mathbf{Edit}\downarrow}$}} \\
\cmidrule(lr){4-5} \cmidrule(lr){6-7} \cmidrule(lr){8-9} \cmidrule(lr){10-11} \cmidrule(lr){12-13} \cmidrule(lr){14-15} 
& & & \textbf{En} & \textbf{Zh} & \textbf{En} & \textbf{Zh} & \textbf{En} & \textbf{Zh} & \textbf{En} & \textbf{Zh} & \textbf{En} & \textbf{Zh} & \textbf{En} & \textbf{Zh} \\
\midrule
\multirow{36}{*}{\rotatebox{90}{\textbf{Expert Models}}} & \multirow{3}{*}{PaddleOCR-VL~\cite{cui2025paddleocr}} &Original& 10.5 & 12.6 & 4.1\phantom{0} & 6.2\phantom{0} & 24.1 & 31.6 & 88.0 & 92.1 & 9.3\phantom{0} & 6.2\phantom{0} & 4.5\phantom{0} & 6.3\phantom{0} \\
& & Photographed & 37.5\loss{27.0} & 39.6\loss{27.0} & 29.4\loss{25.3} & 37.7\loss{31.5} & 46.5\loss{22.4} & 52.6\loss{21.0} & 54.2\loss{33.8} & 65.3\loss{26.8} & 44.4\loss{35.1} & 31.4\loss{25.2} & 28.8\loss{24.3} & 37.9\loss{31.6} \\
& & Unwarping & 15.7\gain{21.8} & 22.0\gain{17.6} & 9.4\phantom{0}\gain{20.0} & 17.6\gain{20.1} & 30.8\gain{15.7} & 41.5\gain{11.1} & 82.9\gain{28.7} & 83.2\gain{17.9} & 13.9\gain{30.5} & 13.5\gain{17.9} & 8.7\phantom{0}\gain{20.1} & 15.4\gain{22.5} \\
\cmidrule{2-15}
& \multirow{3}{*}{MinerU2.5~\cite{niu2025mineru2}} &Original& 11.1 & 17.4 & 5.0\phantom{0} & 7.4\phantom{0} & 25.8 & 47.3 & 88.3 & 89.2 & 8.9\phantom{0} & 8.3\phantom{0} & 4.5\phantom{0} & 6.8\phantom{0} \\
& & Photographed & 37.3\loss{26.2} & 47.4\loss{30.0} & 37.0\loss{32.0} & 53.6\loss{46.2} & 44.3\loss{18.5} & 62.0\loss{14.7} & 54.9\loss{33.4} & 59.8\loss{29.4} & 38.9\loss{30.0} & 33.5\loss{25.2} & 29.0\loss{24.5} & 40.3\loss{33.5} \\
& & Unwarping & 17.3\gain{20.0} & 25.2\gain{22.2} & 13.1\gain{23.9} & 19.1\gain{34.5} & 31.9\gain{12.4} & 52.2\gain{9.8} & 79.2\gain{24.3} & 81.1\gain{21.3} & 15.7\gain{23.2} & 14.6\gain{18.9} & 8.3\phantom{0}\gain{20.7} & 15.0\gain{25.3} \\
\cmidrule{2-15} 
& \multirow{3}{*}{dots.ocr~\cite{dotsocr}} &Original& 12.5 & 16.0 & 3.2\phantom{0} & 6.6\phantom{0} & 32.9 & 41.6 & 88.6 & 89.0 & 9.9\phantom{0} & 9.2\phantom{0} & 4.0\phantom{0} & 6.7\phantom{0} \\
& & Photographed & 33.7\loss{21.2} & 37.3\loss{21.3} & 29.8\loss{26.6} & 35.8\loss{29.2} & 39.2\loss{6.3} & 54.4\loss{12.8} & 63.7\loss{24.9} & 67.6\loss{21.4} & 33.0\loss{23.1} & 27.1\loss{17.9} & 32.8\loss{28.8} & 31.8\loss{25.1} \\
& & Unwarping & 16.3\gain{17.4} & 24.1\gain{13.2} & 8.3\phantom{0}\gain{21.5} & 20.9\gain{14.9} & 32.2\gain{7.0} & 42.0\gain{12.4} & 80.2\gain{16.5} & 82.3\gain{14.7} & 16.9\gain{16.1} & 14.6\gain{12.5} & 7.9\phantom{0}\gain{24.9} & 18.9\gain{12.9} \\
\cmidrule{2-15}
& \multirow{3}{*}{MonkeyOCR~\cite{li2025monkeyocr}} &Original& 14.6 & 22.1 & 6.8\phantom{0} & 11.8 & 27.2 & 45.2  & 81.3 & 85.5 & 14.9 & 13.4 & 9.3\phantom{0} & 17.9 \\
& & Photographed & 46.4\loss{31.8} & 52.8\loss{30.7} & 34.5\loss{27.7} & 43.9\loss{32.1} & 48.7\loss{21.5} & 61.6\loss{16.4} & 33.1\loss{48.2} & 37.4\loss{48.1} & 64.5\loss{49.6} & 61.5\loss{48.1} & 37.9\loss{28.6} & 44.1\loss{26.2} \\
& & Unwarping & 18.8\gain{27.6} & 31.9\gain{20.9} & 12.5\gain{22.0} & 23.6\gain{20.3} & 32.1\gain{16.6} & 55.8\gain{5.8} & 77.2\gain{44.1} & 77.1\gain{39.7} & 17.2\gain{47.3} & 19.5\gain{42.0} & 13.5\gain{24.4} & 28.7\gain{15.4} \\
\cmidrule{2-15}
& \multirow{3}{*}{Dolphin~\cite{feng2025dolphin}} &Original& 20.5 & 31.3 & 9.2\phantom{0} & 20.4 & 44.7 & 60.6 & 76.1 & 66.9 & 19.3 & 28.2 & 8.8\phantom{0} & 11.6 \\
& & Photographed & 57.5\loss{37.0} & 71.5\loss{40.2} & 54.9\loss{45.7} & 71.5\loss{51.1} & 65.6\loss{20.9} & 82.8\loss{22.2} & 33.0\loss{43.1} & 19.3\loss{47.6} & 67.9\loss{48.6} & 73.9\loss{45.7} & 46.2\loss{37.4} & 57.7\loss{46.1} \\
& & Unwarping & 27.3\gain{30.2} & 45.5\gain{26.0} & 17.9\gain{37.0} & 36.9\gain{34.6} & 48.3\gain{17.3} & 75.1\gain{7.7} & 63.8\gain{30.8} & 48.6\gain{29.3} & 29.2\gain{38.7} & 42.5\gain{31.4} & 13.9\gain{32.3} & 27.3\gain{30.4} \\
\cmidrule{2-15}
& \multirow{3}{*}{olmOCR~\cite{poznanski2025olmocr}} &Original& 32.6 & 46.9 & 9.7\phantom{0} & 29.3 & 45.5 & 65.5 & 68.1 & 61.3 & 60.8 & 65.2 & 14.5 & 27.7 \\
& & Photographed & 39.1\loss{6.5} & 46.1\gain{0.8} & 19.3\loss{9.6} & 27.2\gain{2.1} & 50.7\loss{5.2} & 66.9\loss{1.4} & 56.5\loss{11.6} & 56.9\loss{4.4} & 65.6\loss{4.8} & 66.0\loss{0.8} & 20.7\loss{6.2} & 24.4\gain{3.3} \\
& & Unwarping & 31.4\gain{7.7} & 43.1\gain{3.0} & 9.6\phantom{0}\gain{9.7} & 23.7\gain{3.5} & 40.0\gain{10.7} & 61.3\gain{5.6} & 65.8\gain{9.3} & 63.7\gain{6.8} & 62.7\gain{2.9} & 63.3\gain{2.7} & 13.4\gain{7.3} & 23.9\gain{0.5} \\
\cmidrule{2-15}
& \multirow{3}{*}{OCRFlux~\cite{OCRFlux2025}} &Original& 23.8 & 34.9 & 11.2 & 25.6 & 44.7 & 71.6 & 69.0 & 80.0 & 26.9 & 16.2 & 12.6 & 26.3 \\
& & Photographed & 36.2\loss{12.4} & 45.8\loss{10.9} & 30.4\loss{19.2} & 40.4\loss{14.8} & 48.4\loss{3.7} & 81.1\loss{9.5} & 49.5\loss{19.5} & 54.3\loss{25.7} & 29.7\loss{2.8} & 29.7\loss{13.5} & 22.5\loss{9.9} & 32.1\loss{5.8} \\
& & Unwarping & 23.6\gain{12.6} & 37.9\gain{7.9} & 11.8\gain{18.6} & 29.7\gain{10.7} & 42.5\gain{5.9} & 73.7\gain{7.4} & 68.1\gain{18.6} & 72.7\gain{18.4} & 27.6\gain{2.1} & 20.8\gain{8.9} & 12.7\gain{9.8} & 27.3\gain{4.8} \\
\cmidrule{2-15}
& \multirow{3}{*}{SmolDocling~\cite{nassar2025smoldocling}} &Original& 49.3 & 81.6 & 26.2 & 82.8 & 75.3 & 99.7 & 16.5 & 7.3\phantom{0} & 90.8 & 92.7 & 22.7 & 52.2 \\
& & Photographed & 90.1\loss{40.8} & 93.7\loss{12.1} & 89.8\loss{63.6} & 99.2\loss{16.4} & 99.6\loss{24.3} & 99.9\loss{0.2} & 4.4\loss{12.1} & 2.4\loss{4.9} & 98.4\loss{7.6} & 98.8\loss{6.1} & 72.7\loss{50.0} & 75.9\loss{23.7} \\
& & Unwarping & 65.2\gain{24.9} & 92.8\gain{0.9} & 45.6\gain{44.2} & 97.9\gain{1.3} & 92.8\gain{6.8} & 99.7\gain{0.2} & 25.9\gain{21.5} & 1.7\loss{0.7} & 90.0\gain{8.4} & 100.0\loss{1.2} & 38.6\gain{34.1} & 74.6\gain{1.3} \\
\cmidrule{2-15}
& \multirow{3}{*}{Nanonets-OCR~\cite{Nanonets-OCR-S}} &Original& 28.3 & 29.5 & 13.4 & 23.1 & 51.8 & 54.6 &  76.8 & 79.4 & 34.3 & 20.1 & 13.5 & 20.0 \\
& & Photographed & 38.6\loss{10.3} & 52.1\loss{22.6} & 21.0\loss{7.6} & 42.0\loss{18.9} & 48.1\gain{3.7} & 67.0\loss{12.4} & 58.5\loss{18.3} & 50.6\loss{28.8} & 64.1\loss{29.8} & 66.7\loss{46.6} & 21.4\loss{7.9} & 32.7\loss{12.7} \\
& & Unwarping & 32.0\gain{6.6} & 44.4\gain{7.7} & 13.2\gain{7.8} & 30.2\gain{11.8} & 42.6\gain{5.5} & 65.6\gain{1.4} & 59.9\gain{1.4} & 59.8\gain{9.2} & 56.1\gain{8.0} & 56.1\gain{10.6} & 14.4\gain{7.0} & 25.6\gain{7.1} \\
\cmidrule{2-15}
& \multirow{3}{*}{DeepSeek-OCR~\cite{wei2025deepseek}} &Original& 13.4 & 18.1 & 4.6\phantom{0} & 9.7\phantom{0} & 28.5 & 43.3 & 82.6 & 89.0 & 13.8 & 8.8\phantom{0} & 6.7\phantom{0} & 10.5 \\
& & Photographed & 54.4\loss{41.0} & 57.8\loss{39.7} & 56.7\loss{52.1} & 57.6\loss{47.9} & 54.4\loss{25.9} & 74.1\loss{30.8} & 28.0\loss{54.6} & 35.4\loss{53.6} & 64.7\loss{50.9} & 59.2\loss{50.4} & 41.7\loss{35.0} & 40.4\loss{29.9} \\
& & Unwarping & 22.1\gain{32.3} & 33.5\gain{24.3} & 14.9\gain{41.8} & 29.4\gain{28.2} & 32.1\gain{22.3} & 58.8\gain{15.3} & 67.0\gain{39.0} & 75.8\gain{40.4} & 26.7\gain{38.0} & 20.9\gain{38.3} & 14.8\gain{26.9} & 24.9\gain{15.5} \\
\cmidrule{2-15}
& \multirow{3}{*}{olmOCR2~\cite{poznanski2025olmocr}} &Original& 16.1 & 26.7 & 4.8\phantom{0} & 18.5 & 39.2 & 54.3 & 83.7 & 78.5 & 12.3 & 16.5 & 8.1\phantom{0} & 17.4 \\
& & Photographed & 27.8\loss{11.7} & 44.6\loss{17.9} & 22.0\loss{17.2} & 39.9\loss{21.4} & 44.6\loss{5.4} & 74.1\loss{19.8} & 67.6\loss{16.1} & 65.4\loss{13.1} & 24.6\loss{12.3} & 28.5\loss{12.0} & 19.9\loss{11.8} & 36.0\loss{18.6} \\
& & Unwarping & 17.5\gain{10.3} & 37.2\gain{7.4} & 7.3\phantom{0}\gain{14.7} & 32.9\gain{7.0} & 37.5\gain{7.1} & 66.7\gain{7.4} & 81.9\gain{14.3} & 77.2\gain{11.8} & 14.3\gain{10.3} & 19.1\gain{9.4} & 11.0\gain{8.9} & 30.2\gain{5.8} \\
\cmidrule{2-15}
& \multirow{3}{*}{Nanonets-OCR2~\cite{Nanonets-OCR-S}} &Original& 26.6 & 34.9 & 19.4 & 34.3 & 60.0 & 68.0 & 81.5 & 82.5 & 15.5 & 17.9 & 11.6 & 19.4 \\
& & Photographed & 34.2\loss{7.6} & 46.1\loss{11.2} & 25.5\loss{6.1} & 44.6\loss{10.3} & 69.0\loss{9.0} & 76.4\loss{8.4} & 70.7\loss{10.8} & 66.0\loss{16.5} & 22.8\loss{7.3} & 31.9\loss{14.0} & 19.5\loss{7.9} & 31.4\loss{12.0} \\
& & Unwarping & 30.6\gain{3.6} & 40.0\gain{6.1} & 21.1\gain{4.4} & 32.6\gain{12.0} & 65.3\gain{3.7} & 77.3\loss{0.9} & 71.9\gain{1.2} & 73.1\gain{7.1} & 24.8\loss{2.0} & 18.5\gain{13.4} & 17.5\gain{2.0} & 25.2\gain{6.2} \\
\midrule
\multirow{18}{*}{\rotatebox{90}{\textbf{General MLLMs}}} 
& \multirow{3}{*}{Qwen2.5-VL-72B~\cite{bai2025qwen2}} &Original& 21.4 & 26.1 & 9.2\phantom{0} & 18.0 & 31.5 & 43.4 & 82.9 & 83.9 & 34.1 & 26.2 & 10.6 & 16.8 \\
& & Photographed & 41.5\loss{20.1} & 57.0\loss{30.9} & 36.2\loss{27.0} & 56.6\loss{38.6} & 42.2\loss{10.7} & 61.8\loss{18.4} & 57.0\loss{25.9} & 55.5\loss{28.4} & 59.6\loss{25.5} & 58.2\loss{32.0} & 28.1\loss{17.5} & 51.3\loss{34.5} \\
& & Unwarping & 24.0\gain{17.5} & 41.4\gain{15.6} & 11.1\gain{25.1} & 42.7\gain{13.9} & 29.9\gain{12.3} & 48.4\gain{13.4} & 77.4\gain{20.4} & 76.1\gain{20.6} & 42.7\gain{16.9} & 34.9\gain{23.3} & 12.3\gain{15.8} & 39.7\gain{11.6} \\
\cmidrule{2-15}
& \multirow{3}{*}{Gemini2.5-Pro~\cite{comanici2025gemini25pushingfrontier}} &Original& 14.8 & 21.2 & 5.5\phantom{0} & 16.8 & 35.6 & 43.9  & 85.8 & 86.4 & 13.0 & 11.9 & 4.9\phantom{0} & 12.1 \\
& & Photographed & 18.2\loss{3.4} & 30.4\loss{9.2} & 9.8\phantom{0}\loss{4.3} & 27.7\loss{10.9} & 37.1\loss{1.5} & 56.8\loss{12.9} & 81.3\loss{4.5} & 82.9\loss{3.5} & 14.6\loss{1.6} & 13.7\loss{1.8} & 11.2\loss{6.3} & 23.6\loss{11.5} \\
& & Unwarping & 16.9\gain{1.3} & 27.3\gain{3.1} & 9.2\phantom{0}\gain{0.6} & 20.8\gain{6.9} & 35.3\gain{1.8} & 57.0\loss{0.2} & 83.4\gain{2.1} & 85.9\gain{3.0} & 13.1\gain{1.5} & 11.8\gain{1.9} & 10.0\gain{1.2} & 19.8\gain{3.8} \\
\cmidrule{2-15}
& \multirow{3}{*}{Doubao-1.6-v~\cite{guo2025seed1}} &Original& 22.5 & 29.3 & 16.2 & 27.6 & 31.2 & 47.2 & 66.6 & 76.3 & 31.9 & 24.5 & 10.8 & 17.9 \\
& & Photographed & 54.7\loss{32.2} & 55.4\loss{26.1} & 60.6\loss{44.4} & 58.2\loss{30.6} & 51.5\loss{20.3} & 61.1\loss{13.9} & 27.6\loss{39.0} & 37.9\loss{38.4} & 67.0\loss{35.1} & 61.9\loss{37.4} & 39.7\loss{28.9} & 40.2\loss{22.3} \\
& & Unwarping & 30.0\gain{24.7} & 42.5\gain{12.9} & 23.8\gain{36.8} & 41.8\gain{16.4} & 34.5\gain{17.0} & 56.4\gain{4.7} & 55.7\gain{28.1} & 60.8\gain{22.9} & 44.9\gain{22.1} & 42.4\gain{19.5} & 16.7\gain{23.0} & 29.5\gain{10.7} \\
\cmidrule{2-15}
& \multirow{3}{*}{Qwen-VL-Max~\cite{Qwen-VL}} &Original& 16.6 & 26.5 & 5.2\phantom{0} & 20.5 & 32.9 & 44.0 & 84.2 & 86.7 & 22.0 & 23.7 & 6.5\phantom{0} & 17.7 \\
& & Photographed & 27.7\loss{11.1} & 42.7\loss{16.2} & 15.9\loss{10.7} & 41.5\loss{21.0} & 41.8\loss{8.9} & 57.2\loss{13.2} & 71.1\loss{13.1} & 71.6\loss{15.1} & 36.3\loss{14.3} & 38.0\loss{14.3} & 16.8\loss{10.3} & 34.4\loss{16.7} \\
& & Unwarping & 19.0\gain{8.7} & 32.6\gain{10.1} & 6.8\phantom{0}\gain{9.1} & 32.1\gain{9.4} & 33.8\gain{8.0} & 48.5\gain{8.7} & 81.3\gain{10.2} & 83.3\gain{11.7} & 26.5\gain{9.8} & 22.0\gain{16.0} & 9.0\phantom{0}\gain{7.8} & 27.8\gain{6.6} \\
\cmidrule{2-15}
& \multirow{3}{*}{GLM-4.5v~\cite{vteam2025glm45vglm41vthinkingversatilemultimodal}} &Original& 25.5 & 32.0 & 16.1 & 27.7 & 43.8 & 51.8 & 74.0 & 77.4 & 26.9 & 30.5 & 15.4 & 17.9 \\
& & Photographed & 36.7\loss{11.2} & 49.6\loss{17.6} & 26.2\loss{10.1} & 47.7\loss{20.0} & 49.9\loss{6.1} & 66.2\loss{14.4} & 58.9\loss{15.1} & 54.0\loss{23.4} & 43.5\loss{16.6} & 49.0\loss{18.5} & 27.3\loss{11.9} & 35.7\loss{17.8} \\
& & Unwarping & 23.9\gain{12.8} & 36.9\gain{12.7} & 13.1\gain{13.1} & 37.7\gain{10.0} & 39.0\gain{10.9} & 53.5\gain{12.7} & 73.8\gain{14.9} & 75.6\gain{21.6} & 26.9\gain{16.6} & 28.7\gain{20.3} & 16.5\gain{10.8} & 27.7\gain{8.0} \\
\cmidrule{2-15}
& \multirow{3}{*}{Kimi-VL~\cite{team2025kimi}} &Original& 36.5 & 38.7 & 17.2 & 22.0 & 48.6 & 52.2 & 57.1 & 67.8 & 65.9 & 62.5 & 14.3 & 18.1 \\
& & Photographed & 69.6\loss{33.1} & 68.7\loss{30.0} & 66.0\loss{48.8} & 63.5\loss{41.5} & 75.5\loss{26.9} & 82.6\loss{30.4} & 16.4\loss{40.7} & 22.9\loss{44.9} & 85.4\loss{19.5} & 82.2\loss{19.7} & 51.6\loss{37.3} & 46.7\loss{28.6} \\
& & Unwarping & 41.1\gain{28.5} & 50.7\gain{18.0} & 26.3\gain{39.7} & 38.5\gain{25.0} & 50.4\gain{25.1} & 68.8\gain{13.8} & 55.4\gain{39.0} & 62.3\gain{39.4} & 65.4\gain{20.0} & 65.0\gain{17.2} & 22.1\gain{29.5} & 30.7\gain{16.0} \\
\bottomrule
\end{tabular*}
\caption{Comprehensive benchmarking of document parsing performance on DocPTBench. This table presents the full results for all specialized Expert Models and General MLLMs evaluated in our study. It contrasts model accuracy on three document types: pristine Original documents, Photographed images with real-world distortions, and geometrically corrected Unwarping images. The results quantify the performance degradation caused by photographic artifacts (subscript values next to Photographed scores) and the partial recovery achieved through unwarping (subscript values next to Unwarping scores). Lower Edit scores and higher TEDS scores indicate better performance.}
\label{tab:parsing_results_full}
\end{table*}

\begin{table*}[!htbp]
\centering
\scriptsize 
\setlength{\tabcolsep}{4pt} 
\renewcommand{\arraystretch}{1.0} 

\begin{tabular}{c|c|c|llll|llll}
\toprule
\multirow{2}{*}{\textbf{Type}} & \multirow{2}{*}{\textbf{Model}} & \multirow{2}{*}{\textbf{Input}} & \multicolumn{4}{c}{\textbf{En-De}} & \multicolumn{4}{c}{\textbf{Zh-De}} \\
\cmidrule(lr){4-7}  \cmidrule(lr){8-11}
& & & \textbf{BLEU} & \textbf{chrF} & \textbf{METEOR} & \textbf{STEDS} & \textbf{BLEU} & \textbf{chrF} & \textbf{METEOR} & \textbf{STEDS} \\
\midrule
\multirow{12}{*}{\rotatebox{90}{\textbf{Open Source}}} & \multirow{3}{*}{Qwen3-VL-4B~\cite{qwen3vl}} & Text & 54.13 & 75.02 & 65.53 & 95.65 & 35.66 & 62.25 & 49.36 & 93.14 \\
& & Original-Simple & 34.13\loss{20.00} & 57.49\loss{17.53} & 44.46\loss{21.07} & 68.00\loss{27.65} & 17.67\loss{17.99} & 35.65\loss{26.60} & 26.41\loss{22.95} & 68.36\loss{24.78} \\
& & Original-CoT & 17.04\loss{17.09} & 38.92\loss{18.57} & 28.05\loss{16.41} & 47.62\loss{20.38} & 25.51\gain{7.84} & 53.09\gain{17.44} & 37.67\gain{11.26} & 69.04\gain{0.68} \\
\cmidrule{2-11}
& \multirow{3}{*}{Qwen2.5-VL-3B~\cite{bai2025qwen2}} & Text & 47.52 & 69.21 & 56.78 & 87.48 & 27.14 & 52.18 & 37.92 & 84.59 \\
& & Original-Simple & 22.07\loss{25.45} & 44.05\loss{25.16} & 31.22\loss{25.56} & 59.16\loss{28.32} & 10.84\loss{16.30} & 24.98\loss{27.20} & 16.74\loss{21.18} & 54.67\loss{29.92} \\
& & Original-CoT & 17.31\loss{4.76} & 39.35\loss{4.70} & 28.07\loss{3.15} & 48.24\loss{10.92} & 11.13\gain{0.29} & 27.71\gain{2.73} & 19.13\gain{2.39} & 49.18\loss{5.49} \\
\cmidrule{2-11}
& \multirow{3}{*}{InternVL3-2B~\cite{zhu2025internvl3}} & Text & 41.56 & 60.87 & 49.95 & 90.46 & 18.30 & 34.53 & 25.38 & 76.81 \\
& & Original-Simple & 9.12\phantom{0}\loss{32.44} & 24.96\loss{35.91} & 11.90\loss{38.05} & 54.68\loss{35.78} & 4.81\phantom{0}\loss{13.49} & 9.99\phantom{0}\loss{24.54} & 7.86\phantom{0}\loss{17.52} & 54.61\loss{22.20} \\
& & Original-CoT & 13.56\gain{4.44} & 36.42\gain{11.46} & 23.31\gain{11.41} & 52.67\loss{2.01} & 7.67\phantom{0}\gain{2.86} & 26.45\gain{16.46} & 14.70\gain{6.84} & 53.49\loss{1.12} \\
\cmidrule{2-11}
& \multirow{3}{*}{InternVL3.5-2B~\cite{wang2025internvl3}} & Text & 45.06 & 63.01 & 52.78 & 94.67 & 25.61 & 45.05 & 32.77 & 90.00 \\
& & Original-Simple & 15.70\loss{29.36} & 35.23\loss{27.78} & 19.75\loss{33.03} & 63.57\loss{31.10} & 8.11\phantom{0}\loss{17.50} & 12.63\loss{32.42} & 10.44\loss{22.33} & 65.98\loss{24.02} \\
& & Original-CoT & 22.73\gain{7.03} & 48.16\gain{12.93} & 35.05\gain{15.30} & 62.84\loss{0.73} & 12.61\gain{4.50} & 32.29\gain{19.66} & 19.73\gain{9.29} & 61.19\loss{4.79} \\
\midrule
\multirow{12}{*}{\rotatebox{90}{\textbf{Closed Source}}} & \multirow{3}{*}{Qwen-VL-Max~\cite{Qwen-VL}} & Text & 62.48 & 79.68 & 72.10 & 97.15 & 41.32 & 67.34 & 55.28 & 92.54 \\
& & Original-Simple & 40.54\loss{21.94} & 67.77\loss{11.91} & 53.51\loss{18.59} & 70.78\loss{26.37} & 26.03\loss{15.29} & 55.35\loss{11.99} & 38.88\loss{16.40} & 70.89\loss{21.65} \\
& & Original-CoT & 42.03\gain{1.49} & 68.53\gain{0.76} & 55.82\gain{2.31} & 69.11\loss{1.67} & 29.11\gain{3.08} & 59.26\gain{3.91} & 42.71\gain{3.83} & 69.71\loss{1.18} \\
\cmidrule{2-11}
& \multirow{3}{*}{GLM-4.5v~\cite{vteam2025glm45vglm41vthinkingversatilemultimodal}} & Text & 62.75 & 79.95 & 72.27 & 97.61 & 39.12 & 64.94 & 52.96 & 92.36 \\
& & Original-Simple & 41.50\loss{21.25} & 67.88\loss{12.07} & 54.85\loss{17.42} & 73.65\loss{23.96} & 21.11\loss{18.01} & 40.70\loss{24.24} & 30.87\loss{22.09} & 72.63\loss{19.73} \\
& & Original-CoT & 43.63\gain{2.13} & 70.02\gain{2.14} & 57.03\gain{2.18} & 72.31\loss{1.34} & 28.74\gain{7.63} & 58.28\gain{17.58} & 42.42\gain{11.55} & 72.77\gain{0.14} \\
\cmidrule{2-11}
& \multirow{3}{*}{Kimi-VL~\cite{team2025kimi}} & Text & 61.19 & 78.09 & 70.89 & 95.31 & 40.26 & 64.03 & 53.42 & 93.29 \\
& & Original-Simple & 30.31\loss{30.88} & 56.60\loss{21.49} & 42.65\loss{28.24} & 65.67\loss{29.64} & 18.73\loss{21.53} & 42.49\loss{21.54} & 28.14\loss{25.28} & 68.30\loss{24.99} \\
& & Original-CoT & 36.65\gain{6.34} & 63.05\gain{6.45} & 49.89\gain{7.24} & 67.86\gain{2.19} & 26.84\gain{8.11} & 54.11\gain{11.62} & 38.88\gain{10.74} & 69.12\gain{0.82} \\
\cmidrule{2-11}
& \multirow{3}{*}{Doubao-1.6-v~\cite{guo2025seed1}} & Text & 57.42 & 77.76 & 68.20 & 86.37 & 35.36 & 64.60 & 50.25 & 81.72 \\
& & Original-Simple & 39.85\loss{17.57} & 68.59\loss{9.17} & 54.13\loss{14.07} & 69.71\loss{16.66} & 24.73\loss{10.63} & 57.42\loss{7.18} & 39.79\loss{10.46} & 69.97\loss{11.75} \\
& & Original-CoT & 41.24\gain{1.39} & 68.84\gain{0.25} & 54.59\gain{0.46} & 70.79\gain{1.08} & 27.64\gain{2.91} & 59.47\gain{2.05} & 42.54\gain{2.75} & 72.88\gain{2.91} \\
\bottomrule
\end{tabular}
\caption{End-to-end translation performance for En-De and Zh-De on digital-born (Original) documents. This table quantifies the performance gap between a model's pure text translation capability and its end-to-end performance on document images, as well as the improvements yielded by Chain-of-Thought (CoT) prompting. }
\label{tab:translation_result_de}
\end{table*}

\begin{table*}[!htbp]
\centering
\scriptsize 
\setlength{\tabcolsep}{4pt} 
\renewcommand{\arraystretch}{1.0} 

\begin{tabular}{c|c|c|llll|llll}
\toprule
\multirow{2}{*}{\textbf{Type}} & \multirow{2}{*}{\textbf{Model}} & \multirow{2}{*}{\textbf{Input}} & \multicolumn{4}{c}{\textbf{En-Fr}} & \multicolumn{4}{c}{\textbf{Zh-Fr}} \\
\cmidrule(lr){4-7}  \cmidrule(lr){8-11}
& & & \textbf{BLEU} & \textbf{chrF} & \textbf{METEOR} & \textbf{STEDS} & \textbf{BLEU} & \textbf{chrF} & \textbf{METEOR} & \textbf{STEDS} \\
\midrule
\multirow{12}{*}{\rotatebox{90}{\textbf{Open Source}}} & \multirow{3}{*}{Qwen3-VL-4B~\cite{qwen3vl}} & Text & 61.27 & 79.14 & 71.71 & 96.50 & 42.28 & 66.86 & 56.36 & 95.11 \\
& & Original-Simple & 38.87\loss{22.40} & 61.33\loss{17.81} & 49.35\loss{22.36} & 67.58\loss{28.92} & 21.18\loss{21.10} & 38.66\loss{28.20} & 31.15\loss{25.21} & 69.60\loss{25.51} \\
& & Original-CoT & 44.34\gain{5.47} & 66.69\gain{5.36} & 55.75\gain{6.40} & 68.14\gain{0.56} & 32.13\gain{10.95} & 57.92\gain{19.26} & 45.46\gain{14.31} & 70.94\gain{1.34} \\
\cmidrule{2-11}
& \multirow{3}{*}{Qwen2.5-VL-3B~\cite{bai2025qwen2}} & Text & 54.71 & 73.12 & 63.83 & 85.19 & 14.31 & 28.39 & 13.29 & 85.83 \\
& & Original-Simple & 23.38\loss{31.33} & 44.76\loss{28.36} & 30.55\loss{33.28} & 61.06\loss{24.13} & 10.79\loss{3.52} & 21.74\loss{6.65} & 16.93\gain{3.64} & 58.85\loss{26.98} \\
& & Original-CoT & 21.04\loss{2.34} & 41.59\loss{3.17} & 32.33\gain{1.78} & 49.80\loss{11.26} & 12.69\gain{1.90} & 28.67\gain{6.93} & 21.61\gain{4.68} & 49.86\loss{8.99} \\
\cmidrule{2-11}
& \multirow{3}{*}{InternVL3-2B~\cite{zhu2025internvl3}} & Text & 45.13 & 63.23 & 51.56 & 86.69 & 16.37 & 27.19 & 23.39 & 71.20 \\
& & Original-Simple & 9.74\phantom{0}\loss{35.39} & 28.97\loss{34.26} & 11.46\loss{40.10} & 60.10\loss{26.59} & 4.52\phantom{0}\loss{11.85} & 8.08\phantom{0}\loss{19.11} & 6.57\phantom{0}\loss{16.82} & 58.92\loss{12.28} \\
& & Original-CoT & 13.25\gain{3.51} & 35.19\gain{6.22} & 20.56\gain{9.10} & 54.76\loss{5.34} & 10.80\gain{6.28} & 31.86\gain{23.78} & 20.86\gain{14.29} & 56.35\loss{2.57} \\
\cmidrule{2-11}
& \multirow{3}{*}{InternVL3.5-2B~\cite{wang2025internvl3}} & Text & 54.13 & 71.52 & 62.29 & 95.59 & 31.80 & 53.13 & 42.61 & 91.35 \\
& & Original-Simple & 17.27\loss{36.86} & 39.92\loss{31.60} & 21.28\loss{41.01} & 65.69\loss{29.90} & 8.65\phantom{0}\loss{23.15} & 13.44\loss{39.69} & 10.82\loss{31.79} & 67.09\loss{24.26} \\
& & Original-CoT & 28.72\gain{11.45} & 53.51\gain{13.59} & 41.60\gain{20.32} & 64.10\loss{1.59} & 16.89\gain{8.24} & 40.06\gain{26.62} & 27.71\gain{16.89} & 62.40\loss{4.69} \\
\midrule
\multirow{12}{*}{\rotatebox{90}{\textbf{Closed Source}}} & \multirow{3}{*}{Qwen-VL-Max~\cite{Qwen-VL}} & Text & 68.65 & 83.22 & 76.51 & 97.29 & 47.65 & 70.84 & 61.27 & 94.84 \\
& & Original-Simple & 46.02\loss{22.63} & 70.58\loss{12.64} & 58.48\loss{18.03} & 70.37\loss{26.92} & 32.32\loss{15.33} & 59.62\loss{11.22} & 45.97\loss{15.30} & 71.12\loss{23.72} \\
& & Original-CoT & 48.88\gain{2.86} & 72.45\gain{1.87} & 62.17\gain{3.69} & 68.00\loss{2.37} & 37.15\gain{4.83} & 64.11\gain{4.49} & 51.48\gain{5.51} & 70.54\loss{0.58} \\
\cmidrule{2-11}
& \multirow{3}{*}{GLM-4.5v~\cite{vteam2025glm45vglm41vthinkingversatilemultimodal}} & Text & 71.14 & 84.60 & 78.60 & 97.21 & 48.38 & 70.83 & 62.01 & 93.95 \\
& & Original-Simple & 49.69\loss{21.45} & 72.87\loss{11.73} & 61.70\loss{16.90} & 72.90\loss{24.31} & 25.87\loss{22.51} & 43.81\loss{27.02} & 35.56\loss{26.45} & 73.51\loss{20.44} \\
& & Original-CoT & 51.89\gain{2.20} & 75.09\gain{2.22} & 64.49\gain{2.79} & 72.95\gain{0.05} & 35.49\gain{9.62} & 62.43\gain{18.62} & 50.39\gain{14.83} & 72.52\loss{0.99} \\
\cmidrule{2-11}
& \multirow{3}{*}{Kimi-VL~\cite{team2025kimi}} & Text & 71.47 & 84.01 & 79.26 & 97.04 & 49.19 & 70.51 & 63.19 & 96.39 \\
& & Original-Simple & 36.63\loss{34.84} & 60.91\loss{23.10} & 49.31\loss{29.95} & 66.71\loss{30.33} & 24.46\loss{24.73} & 48.70\loss{21.81} & 36.41\loss{26.78} & 68.48\loss{27.91} \\
& & Original-CoT & 45.25\gain{8.62} & 67.89\gain{6.98} & 57.23\gain{7.92} & 68.14\gain{1.43} & 35.28\gain{10.82} & 60.82\gain{12.12} & 48.63\gain{12.22} & 71.01\gain{2.53} \\
\cmidrule{2-11}
& \multirow{3}{*}{Doubao-1.6-v~\cite{guo2025seed1}} & Text & 65.06 & 82.32 & 75.37 & 88.29 & 42.37 & 68.81 & 58.59 & 83.11 \\
& & Original-Simple & 46.28\loss{18.78} & 71.98\loss{10.34} & 59.84\loss{15.53} & 70.87\loss{17.42} & 31.48\loss{10.89} & 62.01\loss{6.80} & 48.59\loss{10.00} & 70.92\loss{12.19} \\
& & Original-CoT & 46.59\gain{0.31} & 71.58\loss{0.40} & 59.95\gain{0.11} & 69.87\loss{1.00} & 34.20\gain{2.72} & 63.63\gain{1.62} & 50.46\gain{1.87} & 72.63\gain{1.71} \\
\bottomrule
\end{tabular}
\caption{End-to-end translation performance for En-Fr and Zh-Fr on digital-born (Original) documents. This table quantifies the performance gap between a model's pure text translation capability and its end-to-end performance on document images, as well as the improvements yielded by Chain-of-Thought (CoT) prompting. }
\label{tab:translation_result_fr}
\end{table*}

\begin{table*}[!htbp]
\centering
\scriptsize 
\setlength{\tabcolsep}{4pt} 
\renewcommand{\arraystretch}{1.0} 

\begin{tabular}{c|c|c|llll|llll}
\toprule
\multirow{2}{*}{\textbf{Type}} & \multirow{2}{*}{\textbf{Model}} & \multirow{2}{*}{\textbf{Input}} & \multicolumn{4}{c}{\textbf{En-Ru}} & \multicolumn{4}{c}{\textbf{Zh-Ru}} \\
\cmidrule(lr){4-7}  \cmidrule(lr){8-11}
& & & \textbf{BLEU} & \textbf{chrF} & \textbf{METEOR} & \textbf{STEDS} & \textbf{BLEU} & \textbf{chrF} & \textbf{METEOR} & \textbf{STEDS} \\
\midrule
\multirow{12}{*}{\rotatebox{90}{\textbf{Open Source}}} & \multirow{3}{*}{Qwen3-VL-4B~\cite{qwen3vl}} & Text & 51.65 & 71.66 & 63.19 & 96.10 & 35.96 & 60.03 & 49.04 & 92.07 \\
& & Original-Simple & 31.78\loss{19.87} & 48.67\loss{22.99} & 40.87\loss{22.32} & 66.66\loss{29.44} & 21.40\loss{14.56} & 42.98\loss{17.05} & 30.63\loss{18.41} & 65.65\loss{26.42} \\
& & Original-CoT & 36.54\gain{4.76} & 58.88\gain{10.21} & 47.84\gain{6.97} & 67.71\gain{1.05} & 26.51\gain{5.11} & 52.07\gain{9.09} & 37.19\gain{6.56} & 68.08\gain{2.43} \\
\cmidrule{2-11}
& \multirow{3}{*}{Qwen2.5-VL-3B~\cite{bai2025qwen2}} & Text & 42.28 & 60.68 & 50.01 & 85.57 & 26.94 & 48.58 & 35.55 & 82.86 \\
& & Original-Simple & 16.23\loss{26.05} & 25.88\loss{34.80} & 20.66\loss{29.35} & 55.28\loss{30.29} & 10.37\loss{16.57} & 23.99\loss{24.59} & 15.53\loss{20.02} & 52.44\loss{30.42} \\
& & Original-CoT & 16.54\gain{0.31} & 35.61\gain{9.73} & 26.27\gain{5.61} & 49.77\loss{5.51} & 10.45\gain{0.08} & 24.57\gain{0.58} & 16.67\gain{1.14} & 47.23\loss{5.21} \\
\cmidrule{2-11}
& \multirow{3}{*}{InternVL3-2B~\cite{zhu2025internvl3}} & Text & 37.09 & 51.06 & 42.39 & 83.92 & 16.57 & 29.06 & 21.80 & 69.09 \\
& & Original-Simple & 8.15\phantom{0}\loss{28.94} & 9.11\phantom{0}\loss{41.95} & 7.23\phantom{0}\loss{35.16} & 59.03\loss{24.89} & 5.05\phantom{0}\loss{11.52} & 6.97\phantom{0}\loss{22.09} & 6.70\phantom{0}\loss{15.10} & 56.33\loss{12.76} \\
& & Original-CoT & 4.67\phantom{0}\loss{3.48} & 11.05\gain{1.94} & 7.99\phantom{0}\gain{0.76} & 46.29\loss{12.74} & 6.95\phantom{0}\gain{1.90} & 22.09\gain{15.12} & 12.11\gain{5.41} & 48.99\loss{7.34} \\
\cmidrule{2-11}
& \multirow{3}{*}{InternVL3.5-2B~\cite{wang2025internvl3}} & Text & 32.47 & 31.08 & 31.68 & 94.63 & 16.93 & 11.17 & 15.32 & 87.96 \\
& & Original-Simple & 13.16\loss{19.31} & 14.73\loss{16.35} & 13.58\loss{18.10} & 64.92\loss{29.71} & 7.11\phantom{0}\loss{9.82} & 8.26\phantom{0}\loss{2.91} & 8.10\phantom{0}\loss{7.22} & 64.16\loss{23.80} \\
& & Original-CoT & 14.01\gain{0.85} & 23.75\gain{9.02} & 18.57\gain{4.99} & 64.22\loss{0.70} & 9.75\phantom{0}\gain{2.64} & 17.98\gain{9.72} & 12.56\gain{4.46} & 59.18\loss{4.98} \\
\midrule
\multirow{12}{*}{\rotatebox{90}{\textbf{Closed Source}}} & \multirow{3}{*}{Qwen-VL-Max~\cite{Qwen-VL}} & Text & 60.59 & 77.25 & 70.04 & 97.25 & 42.64 & 65.64 & 55.19 & 91.62 \\
& & Original-Simple & 36.18\loss{24.41} & 59.93\loss{17.32} & 48.35\loss{21.69} & 70.65\loss{26.60} & 27.47\loss{15.17} & 54.72\loss{10.92} & 39.11\loss{16.08} & 70.32\loss{21.30} \\
& & Original-CoT & 40.08\gain{3.90} & 65.58\gain{5.65} & 53.64\gain{5.29} & 68.33\loss{2.32} & 30.71\gain{3.24} & 58.03\gain{3.31} & 43.49\gain{4.38} & 69.05\loss{1.27} \\
\cmidrule{2-11}
& \multirow{3}{*}{GLM-4.5v~\cite{vteam2025glm45vglm41vthinkingversatilemultimodal}} & Text & 59.71 & 76.51 & 69.41 & 96.08 & 42.45 & 65.67 & 55.38 & 92.19 \\
& & Original-Simple & 37.31\loss{22.40} & 58.54\loss{17.97} & 48.84\loss{20.57} & 72.79\loss{23.29} & 22.05\loss{20.40} & 40.90\loss{24.77} & 30.80\loss{24.58} & 71.26\loss{20.93} \\
& & Original-CoT & 41.45\gain{4.14} & 67.12\gain{8.58} & 55.11\gain{6.27} & 72.51\loss{0.28} & 28.45\gain{6.40} & 56.04\gain{15.14} & 41.36\gain{10.56} & 68.13\loss{3.13} \\
\cmidrule{2-11}
& \multirow{3}{*}{Kimi-VL~\cite{team2025kimi}} & Text & 55.60 & 70.75 & 64.51 & 96.37 & 39.72 & 61.49 & 51.07 & 93.41 \\
& & Original-Simple & 27.07\loss{28.53} & 49.63\loss{21.12} & 37.35\loss{27.16} & 67.48\loss{28.89} & 16.88\loss{22.84} & 38.89\loss{22.60} & 24.15\loss{26.92} & 65.68\loss{27.73} \\
& & Original-CoT & 32.14\gain{5.07} & 57.07\gain{7.44} & 44.33\gain{6.98} & 68.71\gain{1.23} & 25.47\gain{8.59} & 51.19\gain{12.30} & 35.86\gain{11.71} & 68.34\gain{2.66} \\
\cmidrule{2-11}
& \multirow{3}{*}{Doubao-1.6-v~\cite{guo2025seed1}} & Text & 50.35 & 72.64 & 61.69 & 81.70 & 32.57 & 60.88 & 46.74 & 78.10 \\
& & Original-Simple & 34.03\loss{16.32} & 62.95\loss{9.69} & 47.42\loss{14.27} & 68.60\loss{13.10} & 21.82\loss{10.75} & 53.54\loss{7.34} & 35.48\loss{11.26} & 69.17\loss{8.93} \\
& & Original-CoT & 37.63\gain{3.60} & 64.82\gain{1.87} & 51.77\gain{4.35} & 70.06\gain{1.46} & 26.08\gain{4.26} & 55.58\gain{2.04} & 39.82\gain{4.34} & 71.23\gain{2.06} \\
\bottomrule
\end{tabular}
\caption{End-to-end translation performance for En-Ru and Zh-Ru on digital-born (Original) documents. This table quantifies the performance gap between a model's pure text translation capability and its end-to-end performance on document images, as well as the improvements yielded by Chain-of-Thought (CoT) prompting.}
\label{tab:translation_result_ru}
\end{table*}

\end{document}